%% file: main.tex
\documentclass[twocolumn, switch]{article} 

\usepackage{preprint}

\usepackage{amsmath,amsthm,amssymb,amsfonts}

\usepackage{algorithm}
\usepackage{algpseudocode}

\usepackage[numbers,square]{natbib}
\usepackage[utf8]{inputenc}
\usepackage[T1]{fontenc}
\usepackage{textcomp}

\usepackage{graphicx}
\usepackage{float}
\usepackage{array}
\usepackage{booktabs}
\usepackage{multirow}
\usepackage[table]{xcolor}
\usepackage{colortbl}
\usepackage{subfig}
\usepackage{caption}
\usepackage{sidecap}
\sidecaptionvpos{figure}{c}
\usepackage{newfloat}
\DeclareFloatingEnvironment[name={Supplementary Figure}]{suppfigure}

\usepackage{microtype}
\usepackage{nicefrac}
\usepackage[normalem]{ulem}
\usepackage{verbatim}
\usepackage{url}
\usepackage{lineno}
\usepackage{titlesec}

\titlespacing\section{0pt}{12pt plus 3pt minus 3pt}{1pt plus 1pt minus 1pt}
\titlespacing\subsection{0pt}{10pt plus 3pt minus 3pt}{1pt plus 1pt minus 1pt}
\titlespacing\subsubsection{0pt}{8pt plus 3pt minus 3pt}{1pt plus 1pt minus 1pt}

\usepackage{pifont}
\usepackage{tcolorbox}
\usepackage{listings}

\usepackage{xcolor}
\usepackage[
  colorlinks=true,
  linkcolor=purple,
  urlcolor=blue,
  citecolor=cyan,
  anchorcolor=black
]{hyperref}

\usepackage{titling}
\usepackage{footmisc}
\usepackage{orcidlink}
\usepackage{eso-pic}
\usepackage{tikz}

\usepackage{enumitem}

\setlist[itemize]{leftmargin=17pt, itemsep=2pt, topsep=2pt, label=\small$\bullet$}
\titlespacing\subsection{0pt}{6pt}{0pt}
\definecolor{myblue}{RGB}{25,95,173}
\definecolor{myorange}{RGB}{217,55,55}

\useunder{\uline}{\ul}{}

\title{GTA-2: Benchmarking General Tool Agents from Atomic Tool-Use to Open-Ended Workflows}

\newcommand{\Author}[3]{
  \textbf{#1}\textsuperscript{#2}
}

\author{
  \Author{Jize Wang}{1*}{0000-0000-0000-0000} \quad
  \Author{Xuanxuan Liu}{1*}{0000-0000-0000-0000}\quad
  \Author{Yining Li}{2}{0000-0000-0000-0000}\quad
  \Author{Songyang Zhang}{3}{0000-0000-0000-0000} \quad
  \Author{Yijun Wang}{3}{0000-0000-0000-0000} \\
  \Author{Zifei Shan}{3}{0000-0000-0000-0000}\quad
  \Author{Xinyi Le}{1$^\dagger$}{0000-0000-0000-0000} \quad
  \Author{Cailian Chen}{1$^\dagger$}{0000-0000-0000-0000}\quad
  \Author{Xinping Guan}{1$^\dagger$}{0000-0000-0000-0000}\quad
  \Author{Dacheng Tao}{4$^\dagger$}{0000-0000-0000-0000}
}

\date{%
  \textsuperscript{1}Shanghai Jiao Tong University\quad
  \textsuperscript{2}Shanghai AI Laboratory\quad
  \textsuperscript{3}Tencent\\
  \textsuperscript{4}Nanyang Technological University\\[1em]
  \footnotesize \textbf{ Email: } jizewang2000@sjtu.edu.cn \quad *\textbf{ Equal Contribution \quad $^\dagger$ Corresponding Authors} \\
}

\begin{document}

\twocolumn[ 
  \begin{@twocolumnfalse} 

\maketitle
\thispagestyle{empty}

\input{sections/0_abstract}

\vspace{0.35cm}

  \end{@twocolumnfalse} 
] 



\input{sections/1_introduction}

\input{sections/2_related_work}

\input{sections/3_method}
\input{sections/4_evaluation}

\input{sections/5_conclusion}


\small
\bibliography{main}


\footnotesize

\input{sections/6_appendix}

\end{document}

%% file: sections/0_abstract.tex
\begin{abstract}

The development of general-purpose agents requires a shift from executing simple instructions to completing complex, real-world productivity workflows. However, current tool-use benchmarks remain misaligned with real-world requirements, relying on AI-generated queries, dummy tools, and limited system-level coordination.
To address this, we propose GTA-2, a hierarchical benchmark for General Tool Agents (GTA) spanning atomic tool use and open-ended workflows. Built on real-world authenticity, it leverages real user queries, deployed tools, and multimodal contexts. (i) GTA-Atomic, inherited from our prior GTA benchmark, evaluates short-horizon, closed-ended tool-use precision. (ii) GTA-Workflow introduces long-horizon, open-ended tasks for realistic end-to-end completion.
To evaluate open-ended deliverables, we propose a recursive checkpoint-based evaluation mechanism that decomposes objectives into verifiable sub-goals, enabling unified evaluation of both model capabilities and agent execution frameworks (i.e., execution harnesses).
Experiments reveal a pronounced capability cliff: while frontier models already struggle on atomic tasks (below 50\%), they largely fail on workflows, with top models achieving only 14.39\% success. 
Further analysis shows that checkpoint-guided feedback improves performance, while advanced frameworks such as Manus and OpenClaw substantially enhance workflow completion, highlighting the importance of execution harness design beyond the underlying model capacity. These findings provide guidance for developing reliable personal and professional assistants.
Dataset and code will be available at \url{https://github.com/open-compass/GTA}.

\end{abstract}

%% file: sections/1_introduction.tex
\section{Introduction}

The pursuit of general-purpose artificial intelligence has been significantly advanced through the development of agents powered by large language models (LLMs). By integrating planning capabilities with external tools (as seen in frameworks like LangChain~\cite{langchain}, AutoGPT~\cite{autogpt}, and Claude Code~\cite{claudecode}), these agents can now execute diverse tasks ranging from information retrieval~\cite{lightrag},~\cite{hingeabl} to complex content creation~\cite{longcot}, \cite{deepresearch},\cite{scientific}. The core competence of such agents lies in their tool-use proficiency: the ability to reason about user requests, select appropriate tools, and orchestrate their execution to achieve complex objectives. Consequently, evaluating tool-use capabilities has become a foundational challenge.

Initial research in tool agent evaluation primarily focuses on atomic tool-use scenarios. Benchmarks like ToolBench \cite{toolbench} and APIBench \cite{apibench} contribute large-scale API collections for scalable testing. However, a significant gap exists between these evaluations and real-world requirements~\cite{gta}. Many existing benchmarks rely on AI-generated queries that explicitly contain solution steps and tool choices, utilize dummy tools that simulate execution via text, and operate in text-only environments. Such simplifications fail to assess an agent’s genuine problem-solving ability in authentic, multimodal contexts.

To address these limitations in atomic tool-use evaluation, we introduce the GTA (General Tool Agents) benchmark~\cite{gta}. GTA is built upon three pillars of authenticity: (i) real user queries crafted by humans to ensure tool-use requirements and multi-step reasoning; (ii) real deployed tools across perception, operation, logic, and creativity categories, enabling end-to-end task execution; and (iii) real multimodal inputs such as spatial scenes, screenshots, and handwritten materials, closely aligned with practical scenarios. Evaluation on GTA reveals substantial bottlenecks in existing LLMs, with even the most powerful models struggling to complete half of the tasks, underscoring the value of a realistic benchmarking approach.

\vspace{3pt}

Despite the progress, the rapid evolution of LLM agents opens up a new frontier: the automation of complex, long-horizon workflows. Modern applications now involve tasks like writing research reports, planning detailed itineraries, or formulating comprehensive market entry strategies. These workflows are characterized by their extended action sequences, diverse control structures, and high-complexity subtasks that require dynamic planning and robust state tracking. This shift exposes a new critical gap. 
Existing benchmarks, including the original GTA~\cite{gta}, are largely restricted to closed-ended atomic tasks with uniquely determined answers. Furthermore, while benchmarks like SWE-bench \cite{swebench} address specific domains like software engineering, there is still a lack of a universal framework that evaluates agents from basic tool precision to cross-domain workflow mastery.

\begin{table*}[tb!]
  \caption{
  Comparison of benchmarks for LLM-based agent systems. *Real-world means solving the queries is helpful for humans in real life while step-implicit and tool-implicit for LLMs. 
  }
  \label{tab:comparison}
  \centering
  \resizebox{0.99\textwidth}{!}{
  \begin{tabular}{l|ccccccccc}
    \toprule
 \multirow{2}{*}{Method}  & Real-world*  & Real deployed & Multimodal & General  & Long  &  Execution result  & Diagnostic & Agent framework \\
   & user queries &  tools & context inputs & AI assistant  & horizon   & evaluation  & checkpoints & evaluation\\
    \midrule
APIBench \cite{apibench}    &  &  & &   &    &  & &\\
ToolBench \cite{toolbench} &  & \ding{51} & &     &   &   & &\\
APIBank \cite{apibank}    &  &\ding{51} & &  &    &    & &\\
AgentBench \cite{agentbench}    &  & \ding{51} & & \ding{51}  &    & \ding{51} & &\\
m\&m's \cite{mnms}    &   &\ding{51}& \ding{51} &   &  & \ding{51} & &\\
GAIA \cite{gaia}     & \ding{51}  & &  \ding{51}& \ding{51}  &     & \ding{51} & &\\
GAIA-2 \cite{gaia2}    &  \ding{51} & \ding{51} &  & \ding{51}  & \ding{51}   & \ding{51} & &\\
OdysseyBench \cite{odysseybench}    & \ding{51}  &\ding{51}&  & & \ding{51}   & \ding{51}  & &\\
DeepPlanning \cite{deepplanning}    & \ding{51}  &\ding{51}& &   & \ding{51}   & \ding{51} &  \ding{51} &\\
\rowcolor[gray]{.9} \textbf{GTA (Ours)~\cite{gta}}   & \ding{51}  &  \ding{51}& \ding{51}& \ding{51}  &    & \ding{51} & &\\
\rowcolor[gray]{.9} \textbf{GTA-2 (Ours)}   & \ding{51}  &  \ding{51}& \ding{51} & \ding{51}   & \ding{51}  & \ding{51} & \ding{51} & \ding{51}\\
    \bottomrule
  \end{tabular}
}

\vspace{1em} 

  \includegraphics[width=0.99\textwidth]{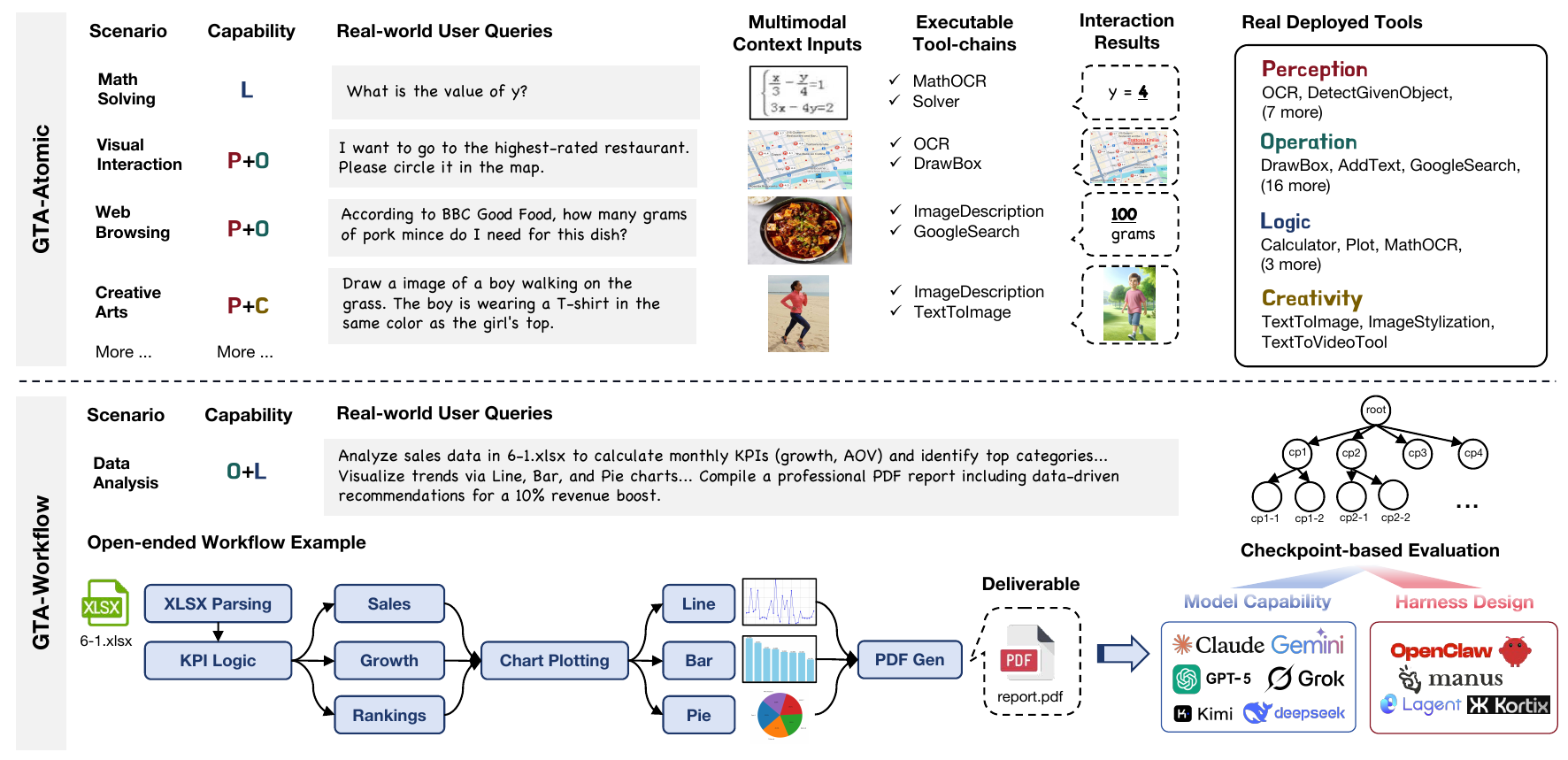}
  \captionof{figure}{The hierarchical framework of GTA-2. \textbf{GTA-Atomic} (Top) evaluates foundational tool-use precision through short-horizon, closed-ended tasks.
\textbf{GTA-Workflow} (Bottom) introduces long-horizon, open-ended productivity tasks.
The benchmark utilizes real user queries, real deployed tools, and multimodal context inputs. For open-ended workflows, a recursive checkpoint-based mechanism is proposed for verifiable evaluation, enabling the assessment of both LLM capabilities and harness design.}
  \label{fig:sample}
\vspace{-10pt}
\end{table*}

\vspace{3pt}
In this paper, we propose GTA-2, a hierarchical benchmark that extends beyond atomic tool use to systematically evaluate long-horizon workflows. GTA-2 consists of two complementary components, as shown in Figure~\ref{fig:sample}: (1) GTA-Atomic, directly inherited from our prior GTA benchmark, which evaluates short-horizon, closed-ended tool-use precision; and (2) GTA-Workflow, a newly introduced and independent evaluation framework designed for long-horizon, open-ended productivity tasks across diverse domains. Importantly, GTA-Workflow is not a simple extension of atomic tasks, but a different setting that targets end-to-end task completion under realistic constraints.

To support this setting, we introduce a recursive checkpoint-based evaluation mechanism that decomposes objectives into verifiable sub-goals. This enables consistent and interpretable evaluation of open-ended deliverables without predefined trajectories. 
GTA-Workflow further provides a unified testbed for evaluating both LLM capabilities and execution frameworks (i.e., execution harnesses), allowing us to analyze how system design affects final outcomes.

Our extensive evaluation reveals a pronounced capability cliff. While frontier models already exhibit limitations in atomic tasks, their performance degrades drastically in workflow settings, with Gemini-2.5-Pro~\cite{gemini2.5} achieving only 14.39\% success rate. Further analysis shows that checkpoint-guided feedback provides moderate improvements, whereas advanced agent frameworks such as Manus and OpenClaw significantly enhance workflow completion. This demonstrates that effective tool orchestration depends not only on model capacity but also critically on execution harness design.

Core contributions of this work are summarized as follows:

\begin{itemize}
\item A hierarchical benchmark for agent evaluation. We propose GTA-2, unifying atomic tool-use and open-ended workflow assessment in a single framework.
\item A novel workflow-centric evaluation paradigm. GTA-Workflow introduces a new, independent benchmark for long-horizon, open-ended tasks, targeting realistic end-to-end productivity scenarios.
\item Checkpoint-based evaluation. We design a mechanism for assessing open-ended deliverables via structured sub-goals, enabling scalable and interpretable evaluation.
\item Joint evaluation of models and agent frameworks. GTA-Workflow serves as a testbed for both LLMs and execution harnesses, revealing the critical role of system design in enabling effective tool use.
\end{itemize}

%% file: sections/2_related_work.tex
\section{Related Work}

\subsection{LLM Agents and Tool Integration}

Recent advances in large language models (LLMs) have enabled agents to interact with external tools for solving complex tasks~\cite{toolsurvey}. Early works such as Toolformer~\cite{toolformer} introduce the paradigm of augmenting language models with autonomous API invocation, while ReAct~\cite{react} formulates tool use as an interleaved process of reasoning and acting. These approaches establish tool use as a core capability of LLM agents, requiring models to interpret user intent, select appropriate tools, and generate executable actions~\cite{apibench}.
Subsequent research has further explored improving tool-use reliability and generalization, including better tool selection, argument prediction, and multi-step decision making~\cite{toolmaker}, \cite{qvalue}. The performance of these systems, however, is intrinsically tied to the underlying LLM's tool-use proficiency, highlighting the critical need for rigorous and realistic benchmarks to quantitatively drive progress in this domain~\cite{berkeley},\cite{metatool}.

\subsection{Agent Execution Frameworks and Harness Design}

Beyond model capability, recent work highlights the critical role of execution frameworks (i.e., agent harnesses) in enabling effective tool use. Early systems such as LangChain~\cite{langchain} and AutoGPT~\cite{autogpt} provide general abstractions for tool integration, but largely rely on fixed execution pipelines. 
Recent advances move toward more structured, system-level designs. Agent Operating Systems~\cite{aios},\cite{memgpt} introduce persistent memory for long-horizon interactions, while runtime systems such as OpenClaw~\cite{openclaw} and MiniMax Agent~\cite{minimaxagent} integrate tool use, memory, and coordination within unified execution environments. 
However, existing studies mainly focus on framework design, leaving the impact of execution harnesses underexplored in standardized evaluation. 
In contrast, GTA-2 provides a unified benchmark to jointly assess LLM capability and harness design, enabling direct measurement of their effect on workflow completion.

\subsection{Long-horizon Agent Workflows}

The community’s focus is rapidly expanding from isolated, atomic tool-use events to the broader challenge of end-to-end workflow management, which is a feature of advanced intelligence. Techniques such as Chain-of-Thought (CoT)~\cite{cot} and Tree of Thoughts (ToT)~\cite{tot} explore structured reasoning and multi-path problem decomposition~\cite{routemoa}, though often in abstract or limited-action settings. Works like Voyager~\cite{voyager} further demonstrate the importance of exploration and state tracking in sequential decision-making within simulated environments.
More recently, there has been a shift toward agents designed for complex, open-ended workflows, such as Claude Code~\cite{claudecode}, Kortix~\cite{suna}, and Manus~\cite{manus}, where success is defined by completing a final deliverable rather than executing a single tool call. These tasks typically involve long action horizons, flexible solution paths, and loosely specified intermediate steps~\cite{travelplanner},\cite{travelsolver}, making them different from closed-ended atomic tasks.

\subsection{Agent Evaluation Benchmarks}

The community has shifted from evaluating isolated tool-use actions to complex agentic sequences~\cite{agentjudge}. Early benchmarks such as ToolBench~\cite{toolbench} and APIBench~\cite{apibench} support large-scale evaluation via extensive APIs, but often rely on synthetic queries or simulated environments, creating a gap with real-world scenarios~\cite{agentcompany},\cite{webarena}. 
To improve realism, high-fidelity benchmarks have emerged in specialized domains. SWE-bench~\cite{swebench} and OSWorld~\cite{osworld} focus on software and OS interaction, while OdysseyBench~\cite{odysseybench} and DeepPlanning~\cite{deepplanning} target long-horizon workflows under constrained settings~\cite{osmarathon}. However, their domain specificity limits coverage of general-purpose, cross-domain workflows. 
Recent efforts such as GAIA-2~\cite{gaia2} expand task diversity, but still rely on simulated execution, leaving a gap for benchmarks combining cross-domain generality with real-world execution. 

Moreover, existing benchmarks primarily assume fixed execution settings, offering limited insight into how agent execution frameworks influence end-to-end task completion. This limitation becomes more pronounced in long-horizon workflows, where execution dynamics play a critical role. In contrast, GTA-2 enables unified evaluation of both LLM capabilities and agent execution frameworks in realistic, open-ended settings.

\begin{figure*}[tb!]
    \centering
    \includegraphics[width=0.9\textwidth]{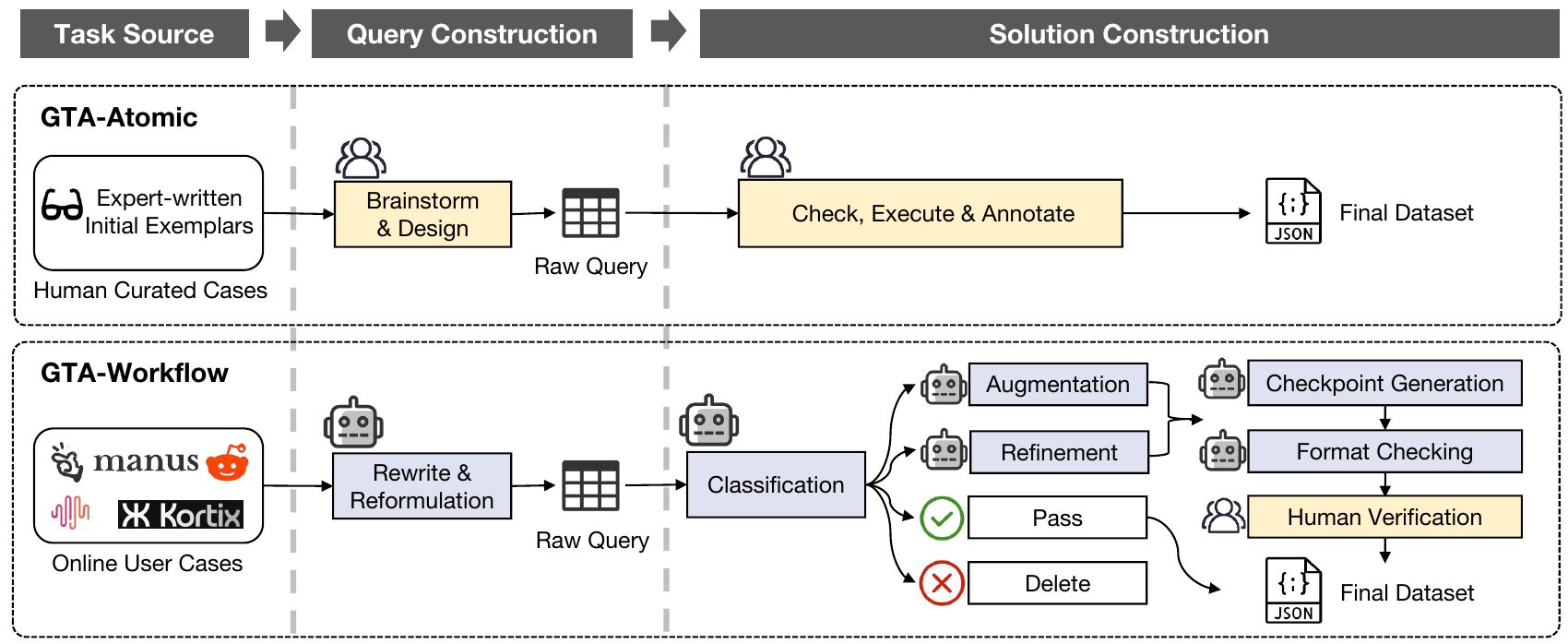}
    \caption{Dataset construction pipeline for the GTA-2 hierarchy. GTA-Atomic (Top): An expert-driven process where initial exemplars are manually expanded to ensure multi-step reasoning precision.
GTA-Workflow (Bottom): A human-in-the-loop semi-automatic pipeline. Tasks are sourced from real-world platforms, then refined and augmented by LLMs with rigorous human verification to guarantee authenticity and feasibility.}
    \label{fig:construct}
\vspace{-15pt}
\end{figure*}

\subsection{Multimodal Interaction}
A competent real-world agent must perceive and act within heterogeneous environments. While earlier Multimodal LLMs (MLLMs) such as GPT-4V~\cite{gpt4v} and LLaVA~\cite{llava} laid the groundwork for visual reasoning, the current frontier, represented by models like GPT-5~\cite{gpt5}, Claude 4.5 Sonnet~\cite{claude4.5}, and Qwen3-VL~\cite{qwen3vl}, demonstrates a much higher level of visual grounding and document understanding. Despite these model-level advances, many agent evaluation frameworks, including the recent GAIA-2~\cite{gaia2}, still primarily rely on text-based interaction or simplified visual abstractions. In actual productivity scenarios, agents must interpret complex visual cues from GUI screenshots~\cite{guigrounding}, parse non-textual information in PDFs~\cite{sail},\cite{docagent}, and maintain spatial awareness across multiple interfaces. Both GTA~\cite{gta} and GTA-2 are architected with multimodal interaction as a first-class citizen, requiring agents to ground their tool-use and workflow execution in authentic, high-fidelity visual and document contexts.

As summarized in Table~\ref{tab:comparison}, despite progress in tool agents, a gap remains for benchmarks that balance realism, generality, and diagnostic depth. The original GTA~\cite{gta} established a high-fidelity testbed for atomic tool use, while GTA-2 extends this paradigm to long-horizon workflows. 
GTA-2 is characterized by four pillars: (1) Authenticity: real-world tools, user queries, and multimodal contexts; (2) Generality: evaluation across diverse, general-purpose tasks in six domains; (3) Atomic-to-Workflow Hierarchy: from precise atomic tool use to open-ended workflow coordination; and (4) Verifiability: a recursive checkpoint-based mechanism for evaluating complex open-ended deliverables. Together, they form a unified framework for assessing general tool agents. 
Moreover, GTA-2 enables unified evaluation of both LLMs and agent execution frameworks, supporting systematic analysis of their impact on end-to-end workflow completion.

%% file: sections/3_method.tex
\section{Hierarchical Design of GTA-2 Benchmark}
This section presents the hierarchical design of GTA-2. We first outline the overall framework, and then focus on the construction and evaluation of GTA-Workflow, which constitutes the primary contribution of this work.

\subsection{Overview}
\label{sec:overview}

The rapid progress of general-purpose agents is driving a transition from solving isolated tool-use problems toward completing complex real-world productivity workflows. Evaluating such agents therefore requires benchmarks that span multiple levels of task complexity while preserving real-world fidelity.
Our prior GTA~\cite{gta} benchmark represents an important step toward realistic evaluation at the atomic-task level, introducing real user queries, executable deployed tools, and multimodal environments. In GTA-2, we directly inherit GTA as \textbf{GTA-Atomic}, which serves as the short-horizon component for evaluating foundational tool-use precision. As its construction protocol remain unchanged, we refer readers to Appendix~\ref{sec:gta1} for details.

To extend evaluation beyond atomic tasks, we introduce \textbf{GTA-Workflow}, a new and independent framework for long-horizon, open-ended productivity tasks. Unlike atomic settings with well-defined objectives, workflow tasks involve flexible solution processes and are evaluated based on final deliverables rather than predefined execution trajectories, posing new challenges for benchmark design and evaluation.
Together, GTA-Atomic and GTA-Workflow form a unified hierarchical framework grounded in real-world authenticity, including real user queries, real deployed tools, and multimodal contexts. This design enables systematic analysis of agent capabilities from precise tool execution to complex workflow completion.

\subsection{Design Principles of GTA-Workflow}

GTA-Workflow is designed as an independent evaluation framework for long-horizon, open-ended productivity tasks. Its design is guided by the following principles:

\textit{Real-world authenticity.}
Consistent with GTA-Atomic, GTA-Workflow is grounded in a shared foundation of real-world authenticity, including (1) \textit{Real User Queries} derived from genuine human needs, (2) \textit{Real Deployed Tools} enabling executable end-to-end interaction, and (3) \textit{Real Multimodal Contexts} requiring agents to operate under realistic inputs. This ensures that workflow tasks more closely reflect practical scenarios rather than synthetic constructions.

\textit{Deliverable-oriented formulation.}
Unlike atomic tasks with predefined answers, GTA-Workflow focuses on long-horizon tasks with explicit deliverables (e.g., reports, code, multimedia artifacts). Evaluation is therefore centered on final output quality rather than intermediate steps, capturing the end-to-end effectiveness of agents in completing real-world objectives. 
We intentionally avoid trajectory-level evaluation in this setting for two reasons. First, valid solution trajectories are inherently diverse and often non-unique in open-ended workflows. Besides, system-level agents typically involve proprietary internal orchestration (e.g., planning, memory), which is not observable or standardized across frameworks. Therefore, trajectory matching is neither a stable nor a fair evaluation objective.

\textit{Goal-driven decomposition with flexible execution.}
To enable structured evaluation of open-ended tasks, each workflow is decomposed into a hierarchy of goal-oriented checkpoints that specify verifiable sub-goals of the final deliverable. These checkpoints describe desired outcomes rather than prescribed actions, allowing agents to adopt diverse execution strategies. 
This design decouples evaluation from specific execution procedures, making it naturally compatible with different agent frameworks. 
As a result, the same task can be consistently evaluated under diverse LLMs and execution systems, enabling systematic analysis of both \textbf{\textit{model capability}} and \textbf{\textit{harness design}} in overall end-to-end performance.

\begin{table}[t]
\centering
\caption{Workflow construction statistics by source. \textit{Initial}: raw tasks; \textit{Final}: tasks retained after processing.}
\label{tab:workflow_construction_source}
\small
\setlength{\tabcolsep}{4pt}
\resizebox{0.49\textwidth}{!}{
\begin{tabular}{lrrrrrr}
\toprule
Source & Initial & Augment & Refine & Delete & Pass & Final \\
\midrule
Manus           & 8  & 3  & 4  & 1  & 0 & 7  \\
Kortix          & 12 & 7  & 5  & 0  & 0 & 12 \\
Flowith         & 6  & 3  & 3  & 0  & 0 & 6  \\
Minimax Agent   & 41 & 21 & 16 & 4  & 0 & 37 \\
CrewAI          & 8  & 2  & 5  & 1  & 0 & 7  \\
Stack Exchange  & 38 & 20 & 16 & 1  & 1 & 37 \\
Reddit          & 41 & 11 & 13 & 15 & 2 & 26 \\
\midrule
Total           & 154 & 67 & 62 & 22 & 3 & 132 \\
\bottomrule
\end{tabular}
}
\end{table}

\begin{table}[t]
\centering
\caption{Magnitude of task rewriting under augmentation and refinement.}
\label{tab:workflow_rewrite_magnitude}
\small
\setlength{\tabcolsep}{2pt}
\resizebox{0.49\textwidth}{!}{
\begin{tabular}{lcccccc}
\toprule
& \multicolumn{2}{c}{Constraints} 
& \multicolumn{2}{c}{Deliverables} 
& \multicolumn{2}{c}{Tools} \\
\cmidrule(lr){2-3} \cmidrule(lr){4-5} \cmidrule(lr){6-7}
Type 
& Avg.\ Added & +(\%) 
& Avg.\ Added & +(\%) 
& Avg.\ Added & +(\%) \\
\midrule
Augment & 3.57 & 73.99 & 1.18 & 493.75 & 3.48 & 208.04 \\
Refine  & 4.45 & 92.31 & 1.81 & 1400.00 & 1.61 & 82.64 \\
\bottomrule
\end{tabular}
}

\end{table}

\subsection{Open-ended Workflow Evaluation}
\label{sec:gta2}

To enable systematic evaluation of open-ended productivity workflows, GTA-Workflow is organized into five components. Section~\ref{sec:workflow_task_source} introduces task sourcing grounded in real-world needs. Section~\ref{sec:workflow_modality} describes the multimodal ecosystem and tool environment. Section~\ref{sec:checkpoint_construct} presents the checkpoint-driven task formulation for structured assessment. 
Section~\ref{sec:workflow_construction} describes the task construction method.
Section~\ref{sec:checkpoint_eval} details the checkpoint-based task evaluation.

\subsubsection{Task Sourcing and Real-World Authenticity}
\label{sec:workflow_task_source}
To align GTA-Workflow with agent development goals and real-world needs, we adopt a dual-source task collection strategy.
\begin{itemize}
\item \textit{\textbf{Agent Platforms.}} We collect cases from platforms including Manus~\cite{manus}, Minimax Agent~\cite{minimaxagent}, Kortix~\cite{suna}, Flowith~\cite{flowith}, and CrewAI~\cite{crewai}, ensuring relevance to current capabilities and practical deployment scenarios.
\item \textit{\textbf{Human Needs.}} We extract and refine high-engagement posts from online communities such as Reddit~\footnote{https://www.reddit.com/} and Stack Exchange~\footnote{https://stackexchange.com/} into benchmark tasks, capturing authentic user demands and increasing workflow diversity.
\end{itemize}

\subsubsection{Multi-Modal Ecosystem and Expanded Tool Set}
\label{sec:workflow_modality}

To evaluate agents on complex workflows, GTA-Workflow adopts a richer and more diverse environment than GTA-Atomic along two dimensions: multimodal inputs and an expanded tool set.
The benchmark supports diverse file types, including images, documents (DOCX, XLSX, PPT, PDF), audio, and video. Compared with GTA-Atomic, which focuses on perception-oriented inputs, GTA-Workflow incorporates broader modalities common in real-world tasks, enabling agents to integrate heterogeneous information for complex deliverables.
The number of tools increases from 14 to 37 (Table~\ref{tab:stat}, Appendix~\ref{appendix:tool}), while retaining the categories of perception, operation, logic, and creation. This expanded set reflects real-world requirements such as audio processing, document editing, and video manipulation, allowing more flexible execution paths without predefined solutions.

\subsubsection{Checkpoint Formulation}
\label{sec:checkpoint_construct}

GTA-Workflow adopts a checkpoint-driven formulation to evaluate open-ended workflows. As execution paths are diverse and outputs are composite artifacts, directly assessing intermediate steps is impractical. Instead, we adopt a deliverable-oriented paradigm, decomposing the user objective into a tree of verifiable sub-goals targeting key aspects of the final deliverable.

First, checkpoints are goal-oriented. Each checkpoint specifies a target state rather than a sequence of actions. For example, it defines \textit{generate an audio clip with a duration between 2.5 and 3.5 minutes} instead of prescribing specific tool calls. This allows flexible execution while keeping evaluation focused on outcome correctness.
Second, checkpoints are organized in a \textit{task → sub-task} hierarchy. Each sub-task defines a verifiable requirement with an associated weight, enabling fine-grained analysis across capabilities. For example, a presentation task may evaluate structure, content coverage, and visual coherence.

\subsubsection{Query and Checkpoint Construction}
\label{sec:workflow_construction}

GTA-Workflow tasks are constructed via a semi-automatic pipeline combining LLM generation with human verification to ensure realism, diversity, and controllability. All prompts related to this section are detailed in Appendix~\ref{appendix:prompt}.

\textit{Initial Task Generation.}
We design structured workflow exemplars specifying task formats, deliverable types, and checkpoint organization. Given raw tasks (Section~\ref{sec:workflow_task_source}), an LLM is prompted with (1) workflow exemplars, (2) the available tool set, and (3) the original user request. The model reformulates each task into a benchmark-ready query with selected tools and a checkpoint tree. This stage produces an initial pool of workflows aligned with tool capabilities.

\textit{Task Refinement and Augmentation.}
Automatically generated tasks may suffer from insufficient complexity, ambiguity, or imbalanced tool usage. We address this through a controlled refinement process, categorizing tasks into \textit{augmentation}, \textit{refinement}, \textit{deletion}, and \textit{pass} via an LLM-based classifier.

\begin{itemize}
\item \textit{Augmentation} increases complexity or expands tool usage by adding requirements.
\item \textit{Refinement} clarifies objectives, constrains output formats (e.g., HTML, PDF), and removes unrealistic content.
\item \textit{Deletion} removes tasks dominated by deep perception requirements beyond the workflow scope.
\item \textit{Pass} retains tasks that meet design criteria.
\end{itemize}

For tasks requiring augmentation or refinement, the LLM rewrites them under explicit guidelines to preserve intent while improving executability and evaluability.

\textit{Task Validation and Checkpoint Regeneration.}
To ensure high-quality task construction, we perform automatic validation on modified queries to enforce key constraints: (i) checkpoints must be outcome-oriented rather than action-oriented, (ii) evaluation criteria must not reference tool invocations, and (iii) task descriptions must avoid predefined execution steps. Queries violating these constraints are iteratively rewritten until compliant. The corresponding checkpoint trees are then regenerated to remain consistent with the updated task.

\textit{Human Verification and Dataset Finalization.}
Finally, human annotators review all tasks to ensure correctness, feasibility, and alignment with GTA-Workflow design principles. Targeted augmentation is applied to underrepresented tools to alleviate long-tail imbalance, with all new tasks undergoing the same refinement and validation process. This human-in-the-loop pipeline ensures high-quality and diverse workflows aligned with real-world scenarios.

\begin{algorithm}[t]
\caption{Recursive Checkpoint Scoring}
\label{alg:recursive_checkpoint_scoring}
\begin{algorithmic}[1]
\Require Checkpoint tree $T$ with root $r$; final deliverable(s) $D$; LLM judge $M$
\Ensure Final score $S(r)\in[0,10]$

\Function{EvalTask}{$T, D, M$}
  \State \Return \Call{ScoreNode}{$r, D, M$}
\EndFunction

\Function{ScoreNode}{$n, D, M$}
  \If{\Call{IsLeaf}{$n$}}
    \State $\mathcal{I} \gets (D,\ \Call{Requirements}{n},\ \Call{Rubric}{n})$
    \State $s \gets \Call{LLMJudge}{M, \mathcal{I}}$ \Comment{$s\in[0,10]$}
    \State \Return $s$
  \EndIf
  \State $\mathcal{C} \gets \Call{Children}{n}$
  \State $\mathbf{w} \gets [\Call{Weight}{c}\mid c\in\mathcal{C}]$
  \State $\mathbf{w} \gets \Call{NormalizeWeights}{\mathbf{w}}$ \Comment{$\sum_{c\in\mathcal{C}} w_c = 1$}
  \State $S \gets 0$
  \ForAll{$c \in \mathcal{C}$}
    \State $s_c \gets \Call{ScoreNode}{c, D, M}$
    \State $S \gets S + \Call{Weight}{c}\cdot s_c$ \Comment{use normalized weights}
  \EndFor
  \State \Return $S$
\EndFunction

\end{algorithmic}
\end{algorithm}

\textit{Task Sources and Rewriting Statistics.}
To improve transparency in workflow construction, Table~\ref{tab:workflow_construction_source} summarizes the source distribution and processing outcomes. 
We collect 154 raw tasks from diverse sources, including agent platforms and online communities, and retain 132 tasks after filtering and rewriting.
We observe that most tasks require non-trivial modification, with 67 tasks undergoing augmentation and 62 tasks requiring refinement, while only 3 tasks directly pass without changes. 
This indicates that raw tasks from real-world sources are often not directly suitable for benchmarking and must be substantially restructured to ensure executability and evaluability.
In terms of source characteristics, tasks from agent platforms exhibit high retention rates, while community-sourced tasks (especially Reddit) show higher deletion rates, reflecting their inherently noisy nature. 
The final dataset maintains a balanced mixture of practical agent workflows and realistic user needs.

Table~\ref{tab:workflow_rewrite_magnitude} further quantifies the magnitude of task rewriting. 
Both augmentation and refinement introduce substantial increases in task complexity across multiple dimensions.
On average, augmentation adds 3.57 constraints, 1.18 deliverable requirements, and 3.48 tools per task, primarily expanding tool coverage and interaction diversity. 
In contrast, refinement introduces larger increases in structural constraints (4.45 on average) and deliverable requirements (1.81 on average), with deliverable-related constraints increasing by up to 1400\%. 
This suggests that refinement mainly strengthens task specification and output requirements, while augmentation focuses on enriching tool usage and task diversity.

\subsubsection{Deliverable-Centric Evaluation}
\label{sec:checkpoint_eval}

GTA-2 adopts a \textit{deliverable-centric} evaluation paradigm, focusing exclusively on final artifacts (e.g., reports, code, multimedia) rather than intermediate execution. The LLM judge does not inspect reasoning or tool usage, but evaluates whether outputs satisfy the goal-oriented requirements defined in the checkpoint tree.
For fine-grained assessment, we use a strong LLM (GPT-5.2) as the judge to score each leaf checkpoint, representing a verifiable sub-goal, on a scale of $[0,10]$ with justification (see Appendix~\ref{appendix:prompt}).
The final task score is computed via recursive aggregation over the checkpoint hierarchy (Algorithm~\ref{alg:recursive_checkpoint_scoring}), where parent nodes are weighted sums of their children. This provides a unified measure of completion while preserving diagnostic granularity. The deliverable-centric design enables fair evaluation based solely on outcome quality.

\begin{figure*}[t]
    \centering
    \includegraphics[width=0.9\textwidth]{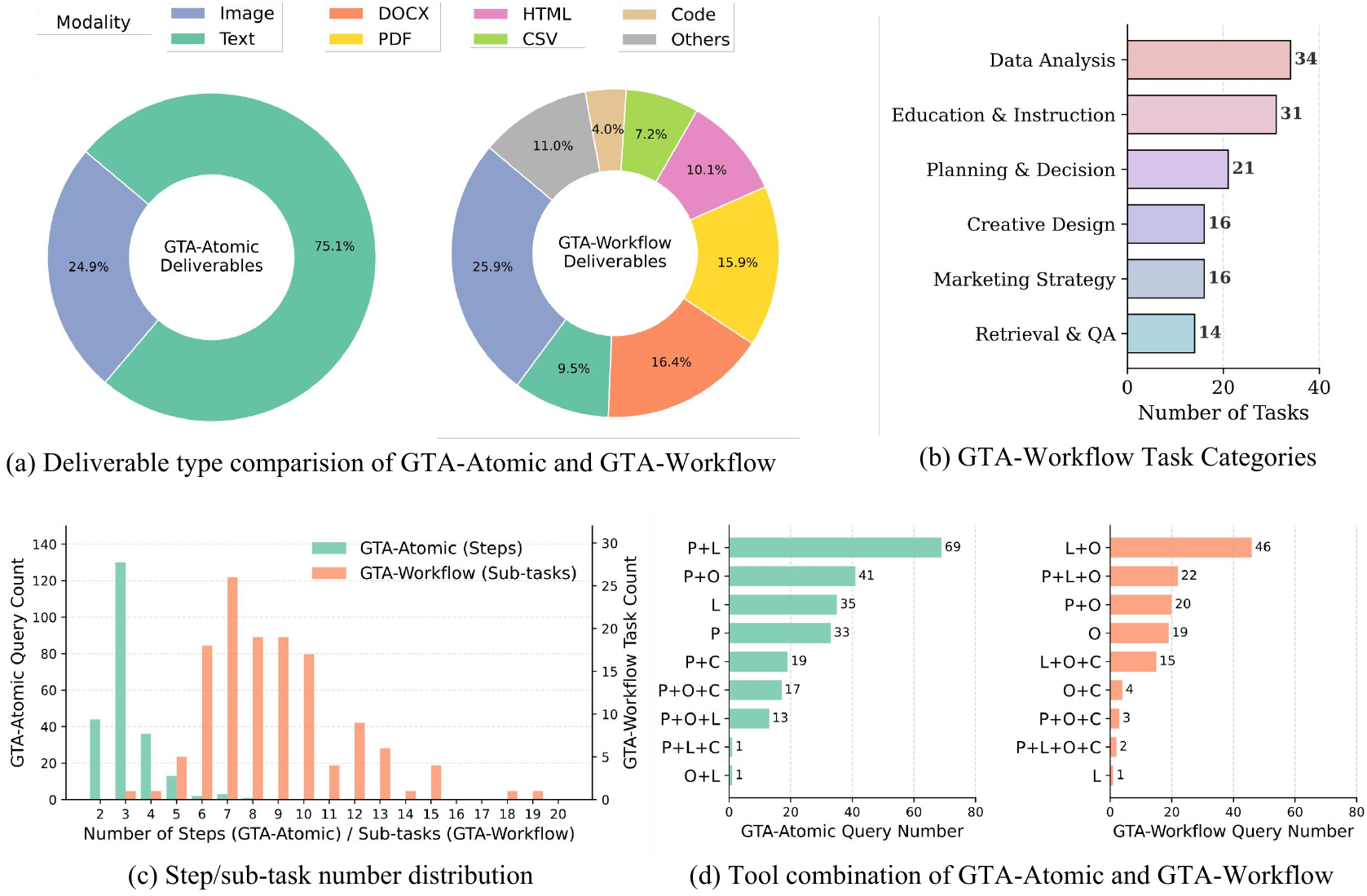}
    \caption{Statistics of GTA-Atomic and GTA-Workflow.}
    \label{fig:statistics}
\vspace{-10pt}
\end{figure*}

\subsection{Dataset Statistics}
\label{sec:dstat}

GTA-2 integrates GTA-Atomic and GTA-Workflow into a hierarchical benchmark spanning structured atomic tool use and open-ended workflow completion. Overall statistics of GTA-2 are summarized in Table~\ref{tab:stat} and Figure~\ref{fig:statistics}.

\textit{GTA-Atomic.}
GTA-Atomic contains 229 tasks with 728 total steps, built upon 14 executable tools, with each task involving 1–4 tools and short tool chains of 2–8 steps. The benchmark is dominated by perception-grounded reasoning patterns (e.g., Perception+Logic), and primarily focuses on structured outputs such as text and images, reflecting its emphasis on step-level tool-use precision.

\textit{GTA-Workflow.}
GTA-Workflow consists of 132 open-ended tasks supported by an expanded set of 37 tools and 1156 sub-tasks. Unlike GTA-Atomic, workflows do not impose fixed execution trajectories, and each task is decomposed into a hierarchy of 3–19 checkpoints representing composite sub-goals. The benchmark significantly broadens both input modalities (e.g., documents, audio, video) and output formats (e.g., reports, code, structured files), covering diverse real-world deliverables across domains. GTA-Workflow is dominated by operation-centered execution (e.g., Logic+Operation), complementing GTA-Atomic.

\begin{table}[tb!]
\centering
    \caption{Statistics of GTA-Atomic and GTA-Workflow. }
    \small
    \resizebox{0.49\textwidth}{!}{%
    \begin{tabular}{l|m{2cm}m{3cm}} 
        \toprule
        Item & GTA-Atomic & GTA-Workflow \\ \midrule
        Total tasks & 229 & 132 \\ 
        Tools & 14 & 37\\ 
        Total steps / sub-tasks & 728 & 1156 \\
        1/2/3/4-tool examples  & 17/147/50/15 & Flexible\\
        \midrule
        Attachment & Image & Image, Audio, Video, PDF, CSV, XLSX \\ 
        \midrule
        Deliverable & Text, Image & Text, Image, Audio, Video, Code, DOCX, PDF, HTML, CSV, PPTX, JSON, Markdown, Other \\
        \bottomrule
    \end{tabular}
}

    \label{tab:stat}
\vspace{-10pt}
\end{table}

%% file: sections/4_evaluation.tex
\section{Experimental Setup}

\subsection{Experiment Settings}

\subsubsection{Models}

For GTA-Atomic, we present 8 representative LLMs to assess foundational tool-use precision in short-horizon settings. Closed-source models include GPT-4~\cite{gpt4}, GPT-4o, Claude-3-Opus~\cite{claude3}, and Mistral-Large~\cite{mistral}. Open-source models cover Llama-3~\cite{llama3}, Mistral~\cite{mistral}, and Mixtral~\cite{mixtral} families.
For GTA-Workflow, we evaluate 13 frontier models to study agent performance in long-horizon scenarios. Closed-source models include GPT-5~\cite{gpt5}, Gemini-2.5-Pro~\cite{gemini2.5}, Claude-Sonnet-4.5~\cite{claude4.5}, Kimi-K2~\cite{kimik2}, Grok-4~\cite{grok4}, DeepSeek-V3.2~\cite{deepseek3.2}, and Llama-4-Scout~\cite{llama4}. Open-source models include Llama-3.1~\cite{llama3} (8B, 70B), Llama-3.2-3B~\cite{llama3}, and Qwen3~\cite{qwen3} (8B, 30B-A3B, 235B-A22B), supporting analysis of model scale, architecture, and reasoning-oriented tuning in open-ended workflows.

\begin{table*}[tb!]
  \caption{\textbf{Main results of GTA-Atomic.} AnsAcc+I denotes AnsAcc w/ ImgGen. P-F1, O-F1, L-F1, C-F1 denote the F1 score of tool selection in Perception, Operation, Logic, and Creativity categories.}
  \label{tab:main_result_atomic}
  \centering
  \resizebox{0.95\textwidth}{!}{
\begin{tabular}{lcccccccccc}
\toprule
\multicolumn{1}{l|}{\multirow{2}{*}{\textbf{Model}}} & \multicolumn{4}{c|}{\textsc{\textbf{Step-by-Step Mode}}}                                                             & \multicolumn{6}{c}{\textsc{\textbf{End-to-End Mode}}} \\
\multicolumn{1}{l|}{}                                & InstAcc                 & ToolAcc                 & ArgAcc                  & \multicolumn{1}{c|}{SummAcc}                 & P-F1    & O-F1   & L-F1 & C-F1  & \textbf{AnsAcc} & \textbf{AnsAcc+I}                  \\ \midrule
\multicolumn{11}{l}{{\cellcolor{cyan!20}\textit{\textbf{Closed-source}}}}                                                                                                                                                         \\

\multicolumn{1}{l|}{GPT-4-1106-Preview}              & 85.19                & 61.40                 &  \textbf{37.88} & \multicolumn{1}{c|}{75.00}                   & 67.61			  & 64.61 & 74.73 & \textbf{89.55} &  \textbf{46.59} &  \textbf{44.90} \\
\multicolumn{1}{l|}{GPT-4o}                          &  \textbf{86.42} & \textbf{70.38} & 35.19                & \multicolumn{1}{c|}{72.77}                & \textbf{75.56}	& \textbf{80.00}&	\textbf{78.75}&	82.35 & 41.52              &40.05 \\
\multicolumn{1}{l|}{Claude-3-Opus}                    & 64.75                & 54.40                 & 17.59                & \multicolumn{1}{c|}{ \textbf{73.81}} & 41.69	&63.23&	46.41&	42.10  & 23.44             & 14.47   \\
\multicolumn{1}{l|}{Mistral-Large}                   & 58.98                & 38.42                & 11.13                & \multicolumn{1}{c|}{68.03}                & 19.17	&30.05&	26.85&	38.89  & 17.06             &11.94  \\ \midrule
\multicolumn{11}{l}{{\cellcolor{green!20}\textit{\textbf{Open-source}}}}                                                                                                                                                     \\

\multicolumn{1}{l|}{Mixtral-8x7B-Instruct}           & 28.67                & 12.03                & 0.36                 & \multicolumn{1}{c|}{54.21}                & 2.19	& 34.69	& 37.68& 	42.55   & 9.77              & 9.33   \\
\multicolumn{1}{l|}{Mistral-7B-Instruct}             & 26.75                & 10.05                & 0.00                    & \multicolumn{1}{c|}{51.06}                & 13.75& 	33.66& 	35.58	& 31.11    & 7.37             &5.54    \\
\multicolumn{1}{l|}{Llama-3-70B-Instruct}             & 47.6                 &  36.80           & 4.31                 & \multicolumn{1}{c|}{ 69.06}         & 32.37	& 22.37& 	36.48& 	31.86    & 8.32              & 6.25   \\
\multicolumn{1}{l|}{Llama-3-8B-Instruct}              &  45.95          & 11.31                & 0.00                    & \multicolumn{1}{c|}{36.88}                & 19.07&	23.23&	29.83&	42.86    & 3.10             &2.74     \\

\bottomrule
\end{tabular}
}

\vspace{1em} 
\setlength\tabcolsep{7pt}
\captionof{table}{\textbf{Main results of GTA-Workflow.} 
SR is short for success rate.
P-SR, O-SR, L-SR, and C-SR denote the Root SR of tasks related to tools in the Perception, Operation, Logic, and Creativity categories, respectively. 
\textbf{Leaf SR} and 
\textbf{Root SR} reflects the fine-grained and coarse-grained overall performance, respectively.}
\vspace{5pt}
  \label{tab:main_result_workflow}
  \centering
  \resizebox{0.85\textwidth}{!}{
\begin{tabular}{lcccccccc}
\toprule
\multicolumn{1}{l|}{\multirow{2}{*}{\textbf{Model}}}            
& \multicolumn{8}{c}{\textsc{\textbf{End-to-End Mode (Deliverable-Centric)}}} \\
\multicolumn{1}{l|}{} &                              Tool SR &  P-SR    & O-SR   & L-SR & C-SR  & Root Score & \textbf{Leaf SR} & \textbf{Root SR}                  \\ \midrule
\multicolumn{9}{l}{{\cellcolor{cyan!20}\textit{\textbf{Closed-source}}}}                                                                                                                                                          \\

\multicolumn{1}{l|}{Gemini-2.5-Pro}              & 	\textbf{91.20}		  & 13.16 & \textbf{13.10} & \textbf{13.93} & \textbf{12.50} & 3.64 & \textbf{28.46} & \textbf{14.39} \\
\multicolumn{1}{l|}{GPT-5}              & 87.31			  & 13.16 & 10.71 & 12.30 & 8.33 & \textbf{3.66} & 26.30 & 11.36 \\

\multicolumn{1}{l|}{Grok-4}              & 87.47			  & 7.89 & 10.71 & 10.66 & 4.17 & 3.56 & 25.17 & 9.85 \\

\multicolumn{1}{l|}{Claude-Sonnet-4.5}              & 88.02			  & 10.53 & 8.33 & 9.84 & 4.17 & 3.50 & 26.21 & 9.09 \\

 \midrule
\multicolumn{9}{l}{{\cellcolor{green!20}\textit{\textbf{Open-source}}}}                                                                                                                                                   \\
\multicolumn{1}{l|}{Qwen3-235B-A22B}              & 88.98			  & \textbf{15.79} & 9.52 & 10.66 & 4.17 & 3.59 & 26.04 & 10.61 \\
\multicolumn{1}{l|}{Llama-4-Scout}              & 87.74			  & \textbf{15.79} & 9.52 & 11.48 & 4.17 & 3.65 & 27.51 & 10.61 \\
\multicolumn{1}{l|}{Deepseek-V3.2}              & 88.81			  & 10.53 & 7.14 & 9.84 & 8.33 & 3.56 & 25.61 & 9.09 \\
\multicolumn{1}{l|}{Kimi-K2}              & 89.85			  & 10.53 & 5.95 & 8.20 & 4.17 & 3.50 & 25.35 & 8.33 \\
\midrule
\multicolumn{1}{l|}{Llama-3.1-70B-Instruct}              & 28.71			  & 2.63 & 1.19 & 0.82 & 0.00 & 1.55 & 3.37 & 0.76 \\
\multicolumn{1}{l|}{Qwen3-30B-A3B}              & 1.94			  & 2.63 & 1.19 & 0.82 & 0.00 & 1.21 & 1.30 & 0.76 \\
\multicolumn{1}{l|}{Llama-3.2-3B-Instruct}              & 0.10			  & 0.00 & 1.19 & 0.82 & 4.17 & 1.02 & 0.78 & 0.76 \\
\multicolumn{1}{l|}{Qwen3-8B}              & 16.97			  & 0.00 & 0.00 & 0.00 & 0.00 & 1.81 & 0.69 & 0.00 \\
\multicolumn{1}{l|}{Llama-3.1-8B-Instruct}              & 13.44			  & 0.00 & 0.00 & 0.00 & 0.00 & 1.18 & 1.47 & 0.00 \\

\bottomrule
\end{tabular}
}

\vspace{1em} 
\setlength\tabcolsep{6pt}
\centering
\captionof{table}{Performance comparison with different agent frameworks (i.e. harness) on a 30-task subset of GTA-Workflow.}
\label{tab:agent_framework}
\vspace{5pt}
\resizebox{0.85\textwidth}{!}{
\begin{tabular}{lcccccc}
\toprule
Harness & Root Score & Leaf SR & Root SR & Total Time & Total Cost & Score/Cost \\
\midrule
\multicolumn{7}{l}{{\cellcolor{cyan!20}\textit{\textbf{Controlled comparison}}}}    \\
Lagent (Claude-Sonnet-4.5) & 2.49 & 10.14  & 0.0  & 50.1 min  & \$10  & 0.249 \\
OpenClaw (Claude-Sonnet-4.5)      & 6.82 & 73.55 & 50.0 & 136.0 min & \$35  & 0.195 \\
\midrule
\multicolumn{7}{l}{{\cellcolor{green!20}\textit{\textbf{System-level comparison}}}}    \\
Manus (Unknown model)                     & 6.94 & 66.67 & 53.3 & 138.6 min & \$15  & 0.463 \\
Kortix (Unknown model)                     & 6.83 & 71.74 & 53.3 & 113.8 min & \$27  & 0.253 \\
\bottomrule
\end{tabular}
}

\end{table*}

\subsubsection{Platform}
\label{sec:platform}
Experiments are conducted on 80GB GPUs using the OpenCompass~\cite{opencompass} evaluation platform. We adopt Lagent~\cite{lagent} as the default agent framework, with ReAct~\cite{react} as the tool invocation schema. Additional details are provided in Appendix~\ref{appendix:agent} and \ref{appendix:prompt}.
Beyond the default setup, we further evaluate advanced agent harnesses including OpenClaw~\cite{openclaw}, Manus~\cite{manus}, and Kortix~\cite{suna}. These frameworks provide structured runtime environments with capabilities such as dynamic planning, persistent memory, and multi-step tool coordination.
We consider two evaluation settings. 
\textit{\textbf{Controlled comparison}} aligns the base model across frameworks to isolate the effect of the execution harness. Specifically, OpenClaw is paired with Claude-Sonnet-4.5 to match the default Lagent configuration. 
\textit{\textbf{System-level comparison}} evaluates closed systems (e.g., Manus and Kortix) under their default configurations, reflecting realistic deployment conditions where model choices are not exposed.
This design allows us to assess harness effectiveness both under controlled conditions and in practical end-to-end systems.

\subsubsection{Evaluation Modes}

For GTA-Atomic, we evaluate models under two complementary modes. \textbf{\textit{Step-by-step mode}} measures fine-grained tool-use precision by requiring the model to predict step $n+1$ given the first $n$ steps, without actual tool execution, enabling direct alignment with ground truth. \textbf{\textit{End-to-end mode}} evaluates dynamic execution, where the model autonomously invokes tools and is assessed based on both tool selection and final outcomes.
For GTA-Workflow, predefined step alignment is impractical due to long-horizon, open-ended tasks. We therefore adopt an \textbf{\textit{end-to-end}} setting with a \textbf{\textit{deliverable-centric}} evaluation. The judge (GPT-5.2) assesses final artifacts against checkpoint requirements (Section~\ref{sec:checkpoint_eval}), emphasizing planning, tool coordination, and outcome quality rather than execution paths.

\subsection{Evaluation Metrics}
\vspace{13pt}
\subsubsection{GTA-Atomic Metrics}
\vspace{10pt}

We design fine-grained metrics covering both tool invocation and execution outcomes. In step-by-step mode, we use four metrics: \textit{\textbf{InstAcc}} (instruction-following accuracy), \textit{\textbf{ToolAcc}} (tool selection accuracy), \textit{\textbf{ArgAcc}} (argument prediction accuracy), and \textit{\textbf{SummAcc}} (final answer summarization accuracy). In end-to-end mode, \textit{\textbf{AnsAcc}} measures overall execution correctness. We also report \textit{\textbf{F1 scores of tool selection}} across perception, operation, logic, and creativity categories.
For AnsAcc, we exclude image generation queries and evaluate only text-based outputs. Objective queries are judged by matching whitelist and blacklist phrases, while subjective queries use cosine similarity between the prediction and three human references, taking the maximum score.
To account for image generation, we introduce \textbf{\textit{AnsAcc w/ ImgGen}}, which evaluates the correctness of predicted generation parameters, as they fully determine the output.

\subsubsection{GTA-Workflow Metrics}

Compared to atomic tasks, GTA-Workflow involves open-ended, long-horizon tasks without unique ground-truth trajectories. We therefore use the \textit{\textbf{Root Score}} ($S_{root} \in [0,10]$) as a fundamental metric. Based on this, we define four metrics to reflect task completion.
(1) \textbf{\textit{Root Success Rate (Root SR).}} 
A task is considered successful if its root score $S_{root} > k$ (default $k=7$). 
Root SR is the proportion of such tasks, reflecting overall task completion.
(2) \textbf{\textit{Leaf Success Rate (Leaf SR).}} 
Leaf SR is the proportion of leaf checkpoints with scores exceeding $k$, measuring fine-grained sub-goal completion.
(3) \textbf{\textit{Tool Success Rate (Tool SR).}} Tool SR measures the percentage of valid tool invocations without system errors or syntax violations, reflecting execution stability.
(4) \textbf{\textit{Capability-Specific SR.}} We report average Root SR across four capability categories: \textit{Perception}, \textit{Operation}, \textit{Logic}, and \textit{Creativity}, capturing problem-solving performance using different capabilities.

\subsubsection{Efficiency Metrics for Harness Evaluation}
In addition to performance metrics, we evaluate the efficiency of different execution frameworks in GTA-Workflow. Specifically, we report:
(1) \textbf{\textit{Total Time}}, the end-to-end execution time for completing a task, reflecting runtime efficiency.
(2) \textbf{\textit{Total Cost}}, the accumulated API cost during task execution.
(3) \textbf{\textit{Score-to-Cost Ratio}}, defined as the achieved Root Score divided by the total cost, measuring the efficiency of converting computational resources into task performance.

\section{Main Results}
\label{sec:gta_main_result}

\subsection{Main Results on GTA-Atomic}

Current LLMs are struggling to accurately invoke tools to solve these real-world tasks. As shown in Table~\ref{tab:main_result_atomic}, the best-performing models, GPT-4 and GPT-4o can only correctly solve fewer than 50\% of the problems, while the rest of the models solve less than 25\%. This shows that real-world problems with implicit steps, real tool invocations, and multimodal inputs impose high requirements on the tool-use capabilities of LLMs.

\subsection{Main Results on GTA-Workflow}

\subsubsection{Model Performance}
The overall performance of representative LLM agents on GTA-Workflow is summarized in Table~\ref{tab:main_result_workflow}. 
A substantial drop is observed in \textit{Root SR} compared to the GTA-Atomic benchmark. 
Even the top-performing model, Gemini-2.5-Pro, achieves an overall root success rate of only 14.39\%, 
despite maintaining a high \textit{Tool SR} (91.20\%). 
This discrepancy suggests that while frontier models have become highly reliable at the atomic level of 
invoking tools correctly, they still struggle to achieve systemic success in long-horizon workflows. 
Completing such workflows requires sustained planning, coordination across multiple tools, and consistent 
state tracking throughout the task. As a result, intermediate errors can propagate 
through the workflow and ultimately lead to the failure of the final deliverable.

\subsubsection{Harness Performance}

We evaluate execution harnesses under two settings introduced in Section~\ref{sec:platform}.

\textit{Controlled comparison.}
With the same base model (Claude-Sonnet-4.5), OpenClaw significantly outperforms the default Lagent setup (Table~\ref{tab:agent_framework}), improving Root Score from 2.49 to 6.82 and Root SR from 0.0\% to 50.0\%.
Leaf SR also increases from 10.14\% to 73.55\%, indicating substantially better sub-goal completion.
These gains can be attributed to the execution harness, demonstrating the importance of structured runtime mechanisms for long-horizon workflows.

\textit{System-level comparison.}
Closed systems such as Manus and Kortix achieve comparable performance (Root Score $\sim$6.8–6.9, Success Rate $>$50\%), reflecting the combined effect of model, harness, and system-level engineering.
These results characterize achievable performance in practical deployments rather than isolated harness effects.
Across both settings, improvements are consistent at the sub-task level, with Leaf SR increasing from 10.14\% (Lagent) to over 65\% in advanced systems, suggesting more stable multi-step execution.

\textit{Efficiency trade-offs.}
Performance gains come with higher cost and latency, as advanced harnesses require longer trajectories and more complex planning.
Among them, Manus achieves the best cost efficiency (Score/Cost 0.463), while OpenClaw prioritizes performance and Kortix provides a balanced trade-off.
These results highlight the critical role of execution harness design in enabling effective long-horizon workflows beyond base model capability.

\begin{table*}[tb!]
\setlength\tabcolsep{8pt}
\centering
\caption{Failure distribution analysis of different LLMs in GTA-Workflow.}
\label{tab:error_distribution}
\small
\resizebox{0.85\textwidth}{!}{
\begin{tabular}{l l | c c c}
\toprule
Stage & Description & Gemini-2.5-Pro & Claude-Sonnet-4.5 & Qwen3-8B \\
\midrule
PLAN     & Requirement-level failures            & 130 (15.7\%) & 173 (20.3\%) & 237 (21.6\%) \\
REASON   & Logical reasoning failures            & 55 (6.7\%)   & 33 (3.9\%)   & 36 (3.3\%)   \\
EXECUTE  & Tool interaction / execution failures & 278 (33.7\%) & 290 (34.0\%) & 250 (22.9\%) \\
GROUNDING& Evidence grounding failures           & 11 (1.3\%)   & 14 (1.6\%)   & 6 (0.5\%)    \\
REFINE   & Post-processing / refinement failures & 101 (12.3\%) & 64 (7.5\%)   & 67 (6.1\%)   \\
ASSEMBLE & Output construction failures          & 15 (1.8\%)   & 60 (7.0\%)   & 58 (5.3\%)   \\
HANDOFF  & Deliverable grounding failures        & 169 (20.4\%) & 148 (17.4\%) & 270 (24.7\%) \\
VERIFY   & Output validation failures            & 7 (0.8\%)    & 7 (0.8\%)    & 8 (0.7\%)    \\
OTHER    & Miscellaneous                         & 61 (7.4\%)   & 64 (7.5\%)   & 162 (14.8\%) \\
\midrule
Total &  & 827 (100\%) & 853 (100\%) & 1094 (100\%) \\
\bottomrule
\end{tabular}
}
\end{table*}

\begin{table}[t]
\centering
\small
\setlength\tabcolsep{3pt}
\caption{Failure distribution of different harnesses.}
\label{tab:harness_error_distribution}
\resizebox{0.495\textwidth}{!}{
\begin{tabular}{l|cccc}
\toprule
Failure Type & Lagent & Openclaw & Manus & Kortix \\
\midrule
Data extraction      & 51 (20.6\%) & 29 (31.5\%) & 20 (27.4\%) & 25 (32.1\%) \\
Reasoning    & 4 (1.6\%)    & 2 (2.2\%)   & 1 (1.4\%) & 3 (3.8\%)\\
Content synthesis          & 73 (29.4\%)    & 21 (22.8\%)   & 17 (23.3\%) & 16 (20.5\%)\\
Formatting     & 120 (48.4\%)    & 40 (43.5\%)   & 35 (47.9\%) & 34 (43.6\%)\\

\midrule
Total & 248 (100\%) & 92 (100\%) & 73 (100\%) & 78 (100\%) \\
\bottomrule
\end{tabular}
}
\end{table}

\subsection{Failure Analysis}

\subsubsection{Model Failure Distribution}
Table~\ref{tab:error_distribution} summarizes stage-wise failure distributions. 
We observe that failures are dominated by execution-stage breakdowns. 
In particular, \textit{EXECUTE} and \textit{HANDOFF} account for the largest proportion across all models, indicating that the primary bottleneck lies in completing end-to-end workflows rather than producing intermediate results.
Specifically, execution failures remain consistently high for frontier models (33.7\% for Gemini-2.5-Pro and 34.0\% for Claude-Sonnet-4.5), reflecting the difficulty of maintaining stable tool interactions over long horizons. 
Meanwhile, deliverable-related failures (\textit{HANDOFF}) are also prominent, especially for smaller models (24.7\% for Qwen3-8B), suggesting that agents frequently fail to produce verifiable final outputs even when partial progress is made.
In contrast, reasoning-related failures (\textit{REASON}) account for a relatively small proportion across all models (3.3\%–6.7\%), indicating that modern LLMs are generally capable of producing locally correct reasoning steps. 
However, these capabilities do not translate into successful task completion, highlighting a gap between local correctness and global execution.
We further observe clear differences across model scales. 
Smaller models exhibit significantly higher planning-stage failures (\textit{PLAN}), while frontier models show a higher proportion of refinement-related failures (\textit{REFINE}), indicating that while they can complete most steps, they often fail to fully satisfy fine-grained quality requirements.
These results reveal that long-horizon workflow failures are primarily driven by execution instability and incomplete deliverable realization, rather than isolated reasoning errors.

\subsubsection{Harness Failure Distribution}

Table~\ref{tab:harness_error_distribution} presents the distribution of failure types across different execution harnesses. 
We observe that formatting-related errors dominate across all methods, accounting for over 40\% of failures in every setting. 
This indicates that even when intermediate steps are correctly executed, agents frequently fail to produce final deliverables that strictly satisfy format and structural requirements.
Compared to the default Lagent setup, advanced harnesses substantially reduce content synthesis failures (from 29.4\% to around 20\%–23\%), suggesting improved capability in organizing and integrating intermediate results. 
However, data extraction failures become relatively more prominent in advanced frameworks, indicating that upstream perception and information retrieval remain non-trivial bottlenecks.
Notably, reasoning errors constitute only a negligible portion of total failures (below 4\% across all methods).
Instead, most failures arise from challenges in execution, integration, and output construction.
This further highlights that long-horizon performance is primarily constrained by system-level execution and output realization, rather than reasoning ability.

\subsubsection{Three-Level Failure Decomposition}

To better understand failure sources beyond process-level analysis, we decompose failures into three levels: leaf-level sub-goal errors (A), mid-level composition errors (B), and final deliverable errors (C). 
These labels are assigned via an LLM-based classifier (GPT-5) operating on checkpoint requirements with standardized guidelines (Appendix~\ref{appendix:prompt}). 
Leaf-level failures (A) correspond to atomic, local sub-goal violations. 
Composition failures (B) occur at the integration level, where sub-goals are largely completed but fail to form a coherent intermediate artifact. 
Notably, leaf-level failures do not necessarily imply composition failures, and composition failures can arise even when most sub-goals are satisfied. 
Deliverable-level failures (C) capture errors in final output realization, including formatting, packaging, file structure, or submission compliance.

\begin{table}[tb!]
\centering
\caption{Three-level failure rates across models and agent systems. A:  leaf-level failure; B: mid-level composition failure; C: final deliverable failure. Lower is better.}
\label{tab:failure_three_level}
\small
\setlength{\tabcolsep}{4pt}
  \resizebox{0.49\textwidth}{!}{
\begin{tabular}{lccc}
\toprule
System & A (Leaf-level) & B (Composition) & C (Deliverable) \\
\midrule
 & Fail Rate $\downarrow$ & Fail Rate $\downarrow$ & Fail Rate $\downarrow$ \\
\multicolumn{4}{l}{\cellcolor{cyan!20}{\textit{\textbf{Models}}}} \\

Gemini-2.5-Pro & 70.09 & 70.83 & 77.78 \\
Claude-Sonnet-4.5 & 72.27 & 70.83 & 80.56 \\
Qwen3-8B & 93.78 & 100.00 & 97.69 \\
\midrule
\multicolumn{4}{l}{\cellcolor{green!20}{\textit{\textbf{Agent Systems}}}} \\
OpenClaw & 31.80 & \textbf{0.00} & 42.59 \\
Manus & \textbf{23.04} & \textbf{0.00} & 42.59 \\
Kortix & 25.81 & 20.00 & \textbf{38.89} \\
\bottomrule
\end{tabular}
}
\end{table}

Results are shown in Table~\ref{tab:failure_three_level}.
Frontier LLMs with default Lagent framework exhibit high failure rates across all levels. 
Deliverable-level failures (C) are most prominent, reaching 77.78\% for Gemini-2.5-Pro and 80.56\% for Claude-Sonnet-4.5. 
This suggests that partially completed sub-goals often fail to translate into correct final outputs. 
Composition-level failures (B) are also high (around 70\%), indicating difficulty in coordinating multiple components.
In contrast, advanced agent systems largely eliminate composition failures. 
Both OpenClaw and Manus achieve 0.00\% failure in B, showing that structured execution harnesses effectively stabilize intermediate composition.
However, deliverable-level failures (C) remain substantial even for these systems (e.g., 42.59\% for OpenClaw and Manus). 
This indicates that output construction and formatting remain key bottlenecks.
Smaller models such as Qwen3-8B fail across all levels, with 100.00\% composition failure, reflecting poor multi-step coherence.
Taken together, failures in long-horizon workflows are mainly driven by composition and deliverable errors rather than atomic sub-goal failures, while execution harnesses are critical for improving intermediate stability.

\begin{figure}[tb!]
    \centering
    \includegraphics[width=0.45\textwidth]{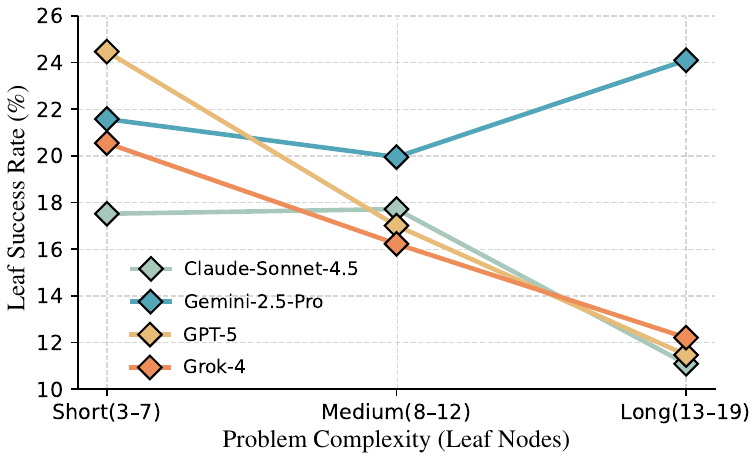}
    \caption{Task difficulty analysis of GTA-Workflow w.r.t. the number of leaf nodes.}
    \label{fig:workflow_complexity}
\end{figure}

\section{Additional Analysis on GTA-Workflow}

\subsection{Scaling and Capability Analysis}

\subsubsection{Performance Gap and Scaling Law}
We observe a clear capability cliff between frontier models and smaller-scale or earlier-generation models. 
Proprietary frontier models, such as Gemini-2.5-Pro and GPT-5, achieve the highest workflow 
completion performance, with success rates ranging from 11.36\% to 14.39\%. Leading open-source models 
(e.g., Qwen3-235B-A22B and Llama-4-Scout) form a second tier with comparable but slightly 
lower root success rates around 10.61\%. 
In contrast, smaller or earlier-generation models exhibit a dramatic performance drop. 
Models such as Llama-3.1-70B-Instruct and Qwen3-30B-A3B achieve success rates below 1\%, 
while smaller models including Qwen3-8B and Llama-3.1-8B-Instruct fail to complete any 
workflow tasks at all (0\% success rate). 
Interestingly, these smaller models can still trigger tool calls successfully 
(13.44\%--16.97\% \textit{Tool Success Rate}), suggesting that correct tool invocation alone is insufficient 
for completing long-horizon workflows.

\subsubsection{Capability-Specific Analysis}

\begin{itemize}
\item \textit{Perception \& Logic:}
Models achieve higher success rates in \textit{Perception} (P.) and \textit{Logic} (L.). Notably, open-source models such as Qwen3-235B-A22B and Llama-4-Scout attain the highest perception success rate (15.79\%), slightly surpassing closed-source models, indicating that recent open-source models have strong multimodal grounding capabilities for 
tool-based perception tasks. 

\item \textit{Operation \& Creativity:}
\textit{Operation} (O.) and \textit{Creativity} (C.) remain the most challenging. Operation tasks require precise interactions with heterogeneous artifacts, increasing execution and state management complexity. Creativity tasks show the lowest success rates, reflecting the difficulty of synthesizing intermediate results into coherent deliverables under multi-stage requirements.

\end{itemize}

\subsection{Task Difficulty Analysis}

\subsubsection{Impact of workflow complexity}
To examine how agent performance scales with task complexity, we evaluate leaf success rate across workflows of increasing size, measured by the number of leaf nodes in the checkpoint tree (Figure~\ref{fig:workflow_complexity}). Most models, including GPT-5, Grok-4, and Claude-Sonnet-4.5, show a clear performance decline as complexity increases from Short (3–7 nodes) to Long (13–19 nodes).
This indicates that maintaining consistent quality becomes more difficult as the number of verifiable sub-tasks grows. An exception is Gemini-2.5-Pro, whose performance slightly drops at the Medium level but recovers at higher complexity, reaching around 24\%. This suggests some frontier models may exhibit stronger robustness in long-horizon tasks, although performance across models converges at high complexity. These results highlight that operational depth and sub-goal coordination remain key bottlenecks for professional-grade agent performance.

\begin{figure}[tb!]
    \centering
    \includegraphics[width=0.49\textwidth]{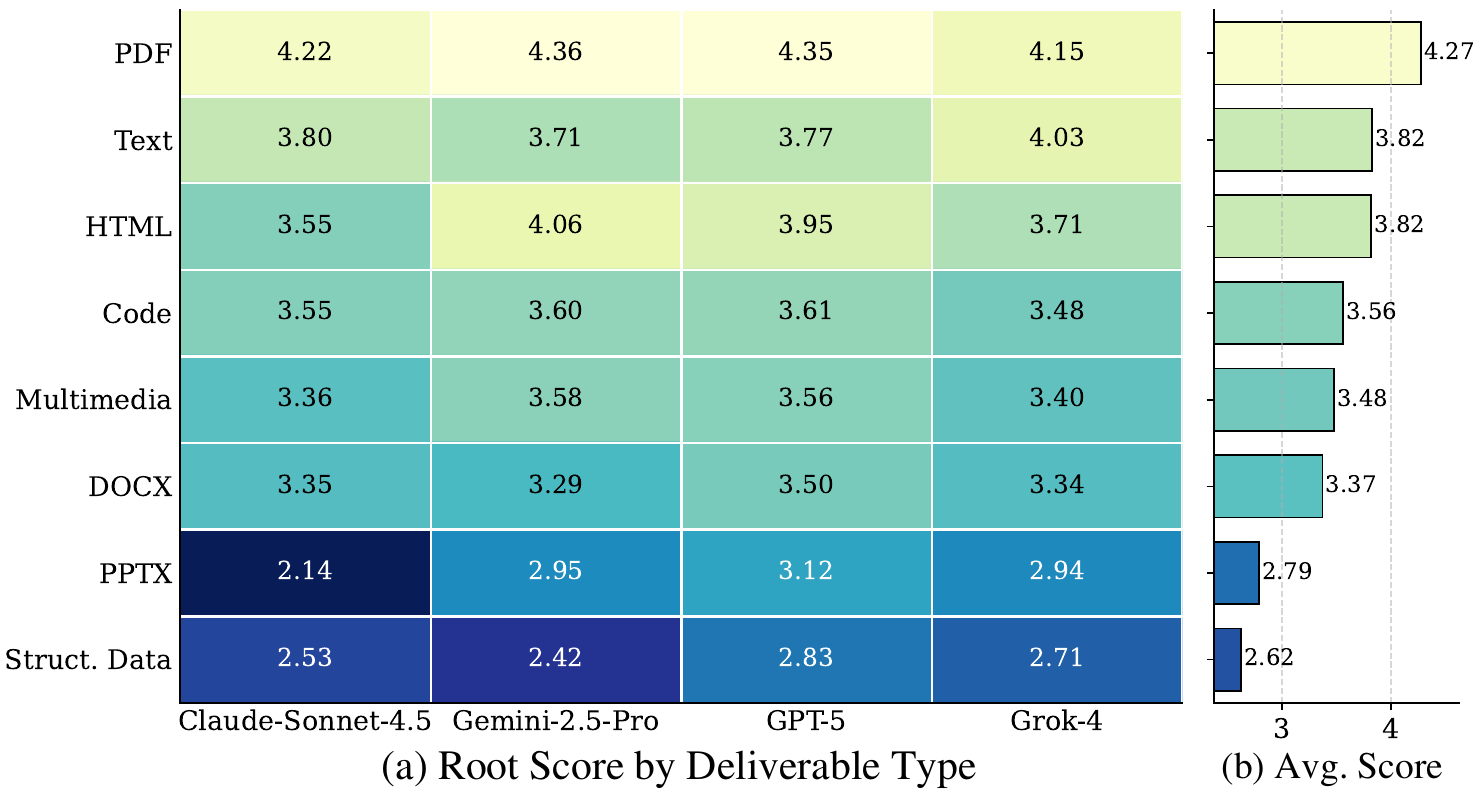}
    \caption{Task difficulty analysis of GTA-Workflow w.r.t. deliverable types. Struct. Data is short for Structured Data, which contains CSV, XLSX, and JSON. Multimedia includes the modalities of image, audio, and video.}
    \label{fig:deliverable_type}
\end{figure}

\begin{figure*}[htbp]
    \centering
    \includegraphics[width= \textwidth]{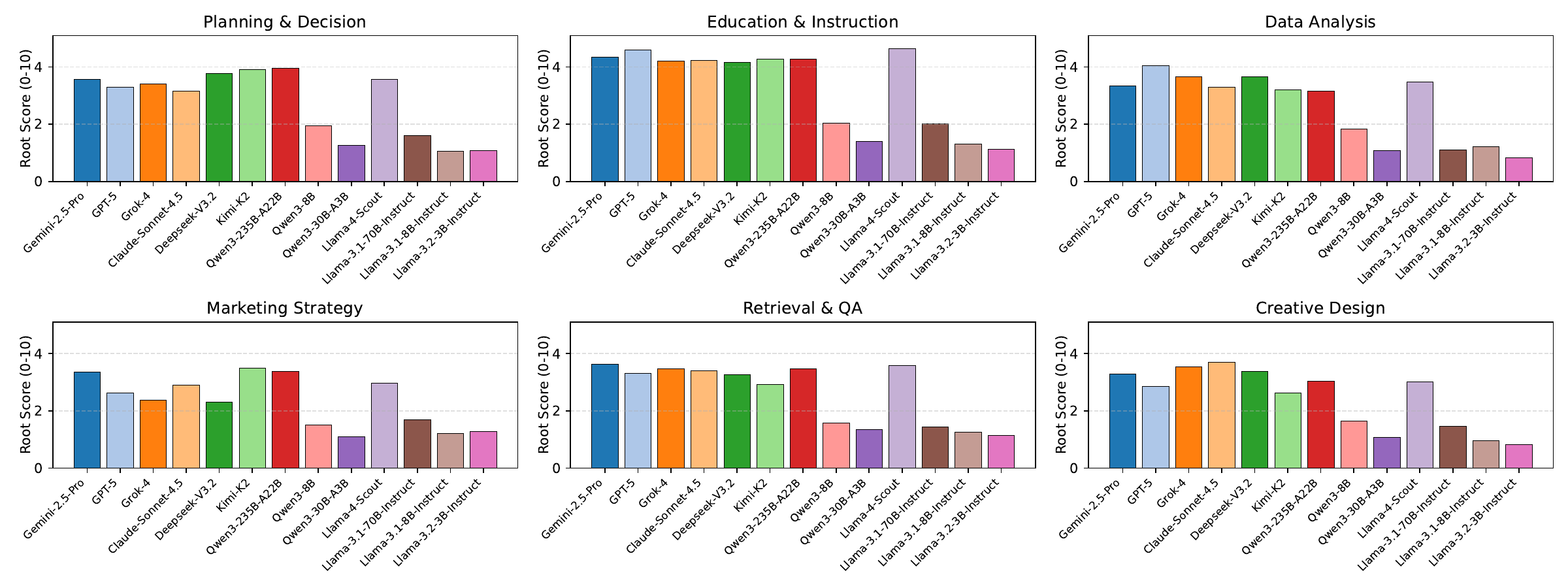}
    \caption{Model performance breakdown across 6 real-world categories in GTA-Workflow. Each subplot displays the average root scores (0-10) calculated via the recursive checkpoint scoring mechanism, with models sorted by performance. The results highlight the varying proficiency of frontier models in handling domain-specific long-horizon workflows.}
    \label{fig:gta2_category}
\end{figure*}

\begin{figure}[tb!]
    \centering
    \includegraphics[width=0.49\textwidth]{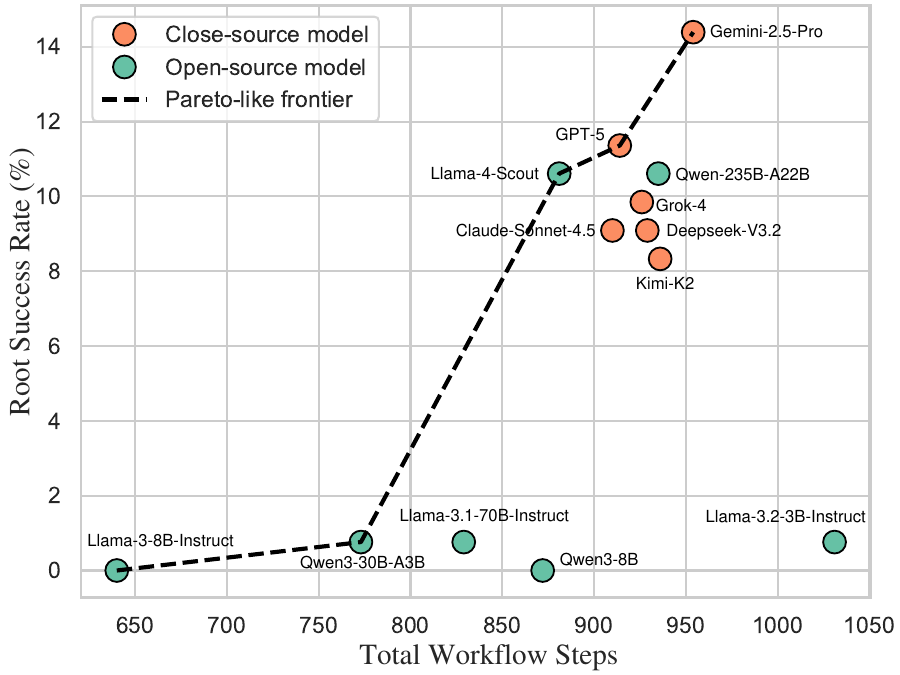}
    \caption{Model efficiency comparison in GTA-Workflow tasks.}
    \label{fig:pareto}
\end{figure}

\subsubsection{Deliverable Type Analysis}
Based on GTA-Workflow, we analyze how different deliverable types affect performance (Figure~\ref{fig:deliverable_type}). Task difficulty is influenced not only by reasoning depth but also by the nature of final artifacts.
Models achieve the best performance on text-based deliverables such as PDF, plain text, and HTML, indicating strong capabilities in long-text synthesis and structured markup. For multimedia outputs (image, audio, video), performance is relatively consistent across models, averaging 3.48, suggesting stable coordination of perception and generation tools.
In contrast, structured data (CSV, XLSX, JSON) and PPTX generation remain challenging, with average scores of 2.62 and 2.79. These tasks require precise logic, cross-file manipulation, and strict schema adherence. A notable gap appears in PPTX tasks, where GPT-5 scores 3.12 while Claude-Sonnet-4.5 scores 2.14, revealing a performance cliff in high-precision data processing.

\subsubsection{Category-Specific Evaluation of GTA-Workflow}

Agent performance varies significantly across workflow domains (Figure~\ref{fig:gta2_category}), and no single model achieves consistent superiority across all categories. For example, Gemini-2.5-Pro leads in Retrieval \& QA, while Claude-Sonnet-4.5 performs slightly better in Creative Design, indicating complementary strengths in reasoning, generation, and multimodal coordination.
A clear capability gap also exists between frontier and non-frontier models. In Education \& Instruction, frontier models consistently score above 4.0, whereas models such as Llama-3.1-70B-Instruct and Qwen3-30B-A3B often score below 2.0 or fail entirely.
Finally, task difficulty varies across domains. Structured reasoning tasks with stable knowledge, such as Education and Knowledge QA, yield higher scores (3.0–3.4), while tasks requiring precise data operations or dynamic interactions, such as Data Analysis and Marketing Strategy, remain more challenging.

\subsection{Tool Execution and Error Analysis}

\subsubsection{Tool Stability vs. Task Completion} Interestingly, a high \textit{Tool SR} does not guarantee task success. For example, Kimi-K2 achieves a high Tool SR of 89.85\%, yet its final \textit{Root SR} is only 8.33\%. This confirms the necessity of our deliverable-centric evaluation: traditional metrics that only monitor whether a tool was called correctly fail to capture the agent's actual problem-solving efficacy. Success in GTA-Workflow requires not just calling the right tools, but using the tools to achieve the goal across a sustained interaction period.

\subsubsection{Model Efficiency Comparison}

To evaluate the trade-off between performance and operational cost in GTA-Workflow, we analyze the relationship between the total workflow steps and the root success rate. As illustrated in Figure~\ref{fig:pareto}, a Pareto-like frontier emerges, representing the optimal balance between efficiency and effectiveness under current technological constraints. Gemini-2.5-Pro occupies the apex of this frontier, achieving the highest Root SR of 14.39\% with relatively moderate step consumption. In contrast, proprietary models like Grok-4 demonstrate relatively high success but involve more redundant steps, whereas open-source models exhibit a sharp divergence. While leading open-source agents like Llama-4-Scout approach the frontier, smaller models often produce a high volume of steps yet fail to achieve any successful outcomes. This disparity suggests that lower-performing models frequently fall into ineffective loops, showing that systemic planning and precision, rather than the number of actions, are the key factors for efficiency in long-horizon tasks.

\subsection{Evaluation Validation}

\subsubsection{Threshold Sensitivity of Success Rate}
\begin{figure}[t]
    \centering
    \includegraphics[width=0.45\textwidth]{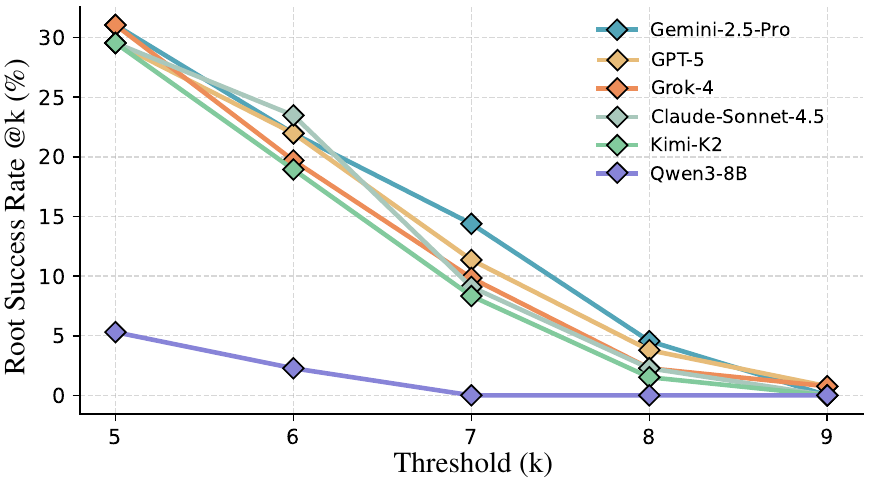}
    \caption{Sensitivity of Root SR to different thresholds.}
    \label{fig:threshold_sensitivity}
\end{figure}

\vspace{-15pt}
We analyze the sensitivity of Root SR under different thresholds by reporting Root SR@k for $k \in \{5,6,7,8,9\}$ across representative models. 
As shown in Figure~\ref{fig:threshold_sensitivity}, the choice of threshold significantly affects both the absolute performance and the discriminability of the metric.
At both low and high thresholds (e.g., $k=5$ and $k=9$), Root SR exhibit clear score clustering, reducing discriminability. In particular, at $k=9$, most models collapse to zero success rate, leading to a loss of informative signal.
Intermediate thresholds such as $k=6$ and $k=8$ partially alleviate this issue but still exhibit noticeable score clustering.
Notably, $k=7$ achieves a better balance: model performance is more evenly distributed, improving discriminability while maintaining a reasonably strict definition of task success. 
We also observe model-specific sensitivity to threshold variation. For example, Claude-Sonnet-4.5 attains the highest score at $k=6$, but drops to near the lowest at $k=7$, indicating that some models tend to produce partially complete outputs that satisfy moderate but not stricter requirements.
Based on these observations, we adopt $k=7$ as the default success threshold, as it provides a balanced trade-off between evaluation strictness and discriminative power. This ensures that the metric remains informative and aligned with the goal of assessing high-quality workflow completion.

\begin{table}[t]
\setlength\tabcolsep{2pt}
\centering
\caption{Agreement between LLM judge and humans. 
}
\resizebox{0.49\textwidth}{!}{
\begin{tabular}{lcccc}
\toprule
Comparison & Pearson & Spearman & ICC(2,1) & MAE \\
\midrule
\multicolumn{5}{l}{\textit{Root Score (30 tasks)}} \\
\midrule
Human1 vs Human2 & 0.965 & 0.873 & 0.949 & 0.603 \\
HumanAvg vs LLM Judge & 0.966 & 0.895 & 0.928 & 0.744 \\
\midrule
\multicolumn{5}{l}{\textit{Leaf Score (276 leaves)}} \\
\midrule
Human1 vs Human2 & 0.901 & 0.757 & 0.883 & 0.794 \\
HumanAvg vs LLM Judge & 0.863 & 0.719 & 0.829 & 1.114 \\
\bottomrule
\end{tabular}
}

\label{tab:judge_agreement}
\end{table}

\begin{table}[t]
\setlength\tabcolsep{6pt}
\centering
\caption{Cross-model validation of LLM judge. Agreement between HumanAvg and LLM judge across outputs from different sources (30 tasks each).}
\label{tab:judge_cross_model}
\resizebox{0.49\textwidth}{!}{
\begin{tabular}{lcccc}
\toprule
Source & Pearson & ICC(2,1) & MAE & Bias \\
\midrule
GPT-5 & 0.966 & 0.928 & 0.744 & +0.639 \\
Gemini-2.5-Pro & 0.937 & 0.890 & 1.006 & +0.930 \\
OpenClaw & 0.941 & 0.910 & 0.613 & +0.460 \\
Qwen3-30B-A3B & 0.926 & 0.848 & 0.561 & +0.561 \\
\bottomrule
\end{tabular}
}
\end{table}

\begin{table}[t]
\setlength\tabcolsep{3pt}
\centering
\caption{Robustness under different judge models.}
\resizebox{0.49\textwidth}{!}{
\begin{tabular}{lcc}
\toprule
Model & GPT-5.2 Judge & Gemini-2.5-Flash Judge \\
\midrule
GPT-5                & 3.66 & 4.69 \\
Gemini-2.5-Pro       & 3.64 & 4.55 \\
Qwen3-235B-A22B      & 3.59 & 4.47 \\
DeepSeek-V3.2        & 3.56 & 4.32 \\
\midrule
\multicolumn{3}{c}{Spearman $\rho$ = 1.00 \quad Kendall $\tau$ = 1.00} \\
\bottomrule
\end{tabular}
}

\label{tab:judge_robustness}
\end{table}

\subsubsection{LLM Judge vs. Human Agreement}
To validate the reliability of automatic evaluation in GTA-Workflow, we conduct a human agreement study on 30 stratified tasks with 276 leaf checkpoints. Outputs from GPT-5 are independently evaluated by two annotators using the same checkpoint tree, and compared with LLM judge scores.
As shown in Table~\ref{tab:judge_agreement}, the LLM judge achieves strong agreement with human evaluation at the task level, with Pearson 0.966 and ICC 0.928. The mean absolute error is 0.74 on a 10-point scale, and 76.7\% of tasks differ by at most one point, comparable to human inter-annotator agreement (Pearson 0.965, ICC 0.949).
At the checkpoint level, agreement remains high but slightly lower, with Pearson 0.863 and ICC 0.829. Converting to pass/fail yields Cohen’s $\kappa$ of 0.812 and 95.3\% accuracy. Aggregated leaf pass rates also correlate strongly with human scores (Pearson 0.952), with an average difference of 4.1\%.
Overall, the LLM judge provides a reliable approximation of human evaluation, achieving near-human consistency while enabling scalable assessment.

To further evaluate the robustness of the LLM judge across different output distributions, we conduct an additional cross-model validation study. 
Specifically, we sample 30 tasks each from four representative sources, including GPT-5, Gemini-2.5-Pro, OpenClaw, and a weaker model (Qwen3-30B-A3B), and compare LLM judge scores with human annotations.
As shown in Table~\ref{tab:judge_cross_model}, strong agreement is consistently observed across all sources. 
Root-level Pearson correlations remain above 0.92 for all models, and ICC scores are consistently high (0.85–0.93), indicating stable agreement between the LLM judge and human evaluation regardless of output origin.

\subsubsection{Judge Robustness}

We replace the original judge (GPT-5.2) with Gemini-2.5-Flash and re-evaluate LLMs. As shown in Table~\ref{tab:judge_robustness}, although Gemini-2.5-Flash produces consistently higher absolute scores, the relative ranking remains identical. This is further confirmed by perfect rank correlation (Spearman $\rho = 1.0$, Kendall $\tau = 1.0$). 
These results indicate that our evaluation framework is robust to the choice of LLM judge: while different judges may exhibit calibration differences, they generally yield consistent comparative conclusions.

\subsubsection{Evaluation Cost Analysis}

We estimate the cost of LLM-based evaluation in GTA-Workflow (Table~\ref{tab:eval_cost}). 
Using GPT-5.2 as the judge, evaluating all 132 tasks under the default Lagent setup costs approximately \$5. 
In contrast, evaluating advanced harnesses (e.g., OpenClaw, Manus) costs around \$10 on a 30-task subset, due to more complex outputs. 
Overall, the evaluation cost remains moderate and scales with task complexity, demonstrating the practicality of our framework for large-scale benchmarking.

\begin{table}[t]
\setlength\tabcolsep{8pt}
\centering
\caption{Evaluation cost using GPT-5.2 as the judge.}
\label{tab:eval_cost}
\resizebox{0.49\textwidth}{!}{
\small
\begin{tabular}{l c c}
\toprule
Setting & \#Tasks & Cost \\
\midrule
Default (Lagent) & 132 & $\sim$ \$5 \\
Advanced Harness (OpenClaw/Manus) & 30 & $\sim$ \$10 \\
\bottomrule
\end{tabular}
}
\end{table}

\begin{table}[t]
\setlength\tabcolsep{2pt}
\centering
\caption{Performance improvement with different feedback types on GTA-Workflow.}
\resizebox{0.49\textwidth}{!}{
\small
\begin{tabular}{lcccc}
\toprule
Setting & Root Score & Improvement & Relative Gain \\
\midrule
GPT-5 & 2.83 & -- & --  \\
+ Coarse Feedback & 2.93 & +0.11 & +4.05\%  \\
+ Checkpoint Feedback & \textbf{3.15} & \textbf{+0.34} & \textbf{+12.03\%}  \\
\bottomrule
\end{tabular}
}

\label{tab:feedback_improvement}
\end{table}

\subsection{Improving Performance}

We further examine whether the checkpoint-based evaluation mechanism in GTA-Workflow can provide effective feedback signals for iterative agent improvement. 
Experiments are conducted on the same stratified subset of 30 workflow tasks used in the validation study. 
For each task, the agent (GPT-5) first generates an initial deliverable, which is then evaluated by the LLM judge using the checkpoint tree. 
The agent performs a second attempt with additional feedback appended to the prompt. 
We compare two settings: \textbf{\textit{coarse feedback}}, which only indicates that the result is incorrect without any cause, and \textbf{\textit{checkpoint feedback}}, which returns detailed failure diagnostics derived from the checkpoint evaluation.

Table~\ref{tab:feedback_improvement} shows the results. 
The initial attempt achieves an average task score of 2.83. 
Providing coarse feedback yields a small improvement to 2.93 (+4.05\%), indicating that even generic retry instructions can slightly improve performance. 
In contrast, checkpoint feedback increases the score to 3.15, corresponding to a 12.03\% improvement over the initial attempt and a 7.66\% gain over coarse feedback. 
These results demonstrate that while general feedback provides limited benefits, fine-grained checkpoint diagnostics offer a more effective signal for correcting errors and refining deliverables. 
This suggests that the recursive checkpoint mechanism in GTA-Workflow can serve not only as an evaluation framework but also as a practical tool for iterative agent optimization.

%% file: sections/5_conclusion.tex
\section{Conclusion}

In conclusion, we present GTA-2, a hierarchical benchmark spanning atomic tasks and long-horizon workflows. At its core, GTA-Workflow introduces a deliverable-centric paradigm for evaluating open-ended, real-world productivity tasks under realistic tool and multimodal settings. We further propose a checkpoint-based evaluation framework to systematically assess complex deliverables. Experiments reveal a significant capability gap, with frontier models achieving only 14.39\% root success rate on workflows. Notably, advanced execution frameworks (e.g., Manus and OpenClaw) substantially improve performance, highlighting the critical role of execution harness design beyond model capability. Overall, GTA-2 provides a rigorous testbed for advancing reliable, professional autonomous agents.

\textbf{Limitations and future work.} 
Despite its strengths, GTA-2 has several limitations. 
GTA-2 is designed as a high-fidelity capability benchmark for realistic workflows, rather than a complete characterization of real-world workflow distributions, isolated harness causality, deployment safety, or the full causal structure of workflow failures.
First, workflow tasks are partially constructed via LLM-based reformulation of real-world cases, which may introduce benchmark construction bias in task formulation and checkpoint design. 
Second, harness comparisons involve both controlled and system-level settings: while comparisons such as Lagent vs. OpenClaw isolate the effect of harness design under a fixed base model, evaluations of closed systems (e.g., Manus and Kortix) reflect the combined contributions of model, harness, and product-level engineering. 
Therefore, conclusions about harness effectiveness should be interpreted at two levels: controlled evidence for harness effects and practical system-level performance. 
Third, the deliverable-centric evaluation focuses on outcome quality and does not explicitly assess safety or deployment readiness; high scores do not necessarily imply safe or reliable real-world deployment. 
Important aspects such as safety, authority control, privacy protection, and governance remain complementary dimensions beyond the current scope. 
Finally, the current failure taxonomy is approximate and partially heuristic, and the mapping between stage-wise labels and final outcome errors is not strictly orthogonal.

Future work will focus on addressing these limitations along several directions. 
First, to reduce benchmark construction bias, we plan to release source-level data and provide paired raw and reformulated examples for greater transparency. 
Second, to better isolate harness effects, we aim to expand controlled comparisons across diverse models and execution frameworks. 
Third, we will incorporate safety-oriented evaluation dimensions, including robustness, authority control, and privacy considerations. 
Finally, we seek to move beyond heuristic failure taxonomies toward more principled causal modeling of workflow failures.

%% file: sections/6_appendix.tex
\clearpage
\appendix
\section*{Additional GTA-2 Information}
\subsection{Tool Definition}
\label{appendix:tool}
The detailed definition of 14 tools across perception, operation, logic, and creativity categories in GTA-Atomic are shown in Table~\ref{tab:tool_desc}.
The detailed definition of extended tools for GTA-Workflow are shown in Table~\ref{tab:ex_tool_desc}.

\subsection{Build an LLM-Based Agent System}
\label{appendix:agent}
We build the LLM-based agent system using Lagent \footnote{ \url{https://github.com/InternLM/lagent}}  framework. It equips an LLM with some action \& planning schema, using action executor to let it interact with external tools. To build such an agent system, we should consider three parts: LLM, action \& planning schema, and tools. In our experiment, we use ReAct as the action \& planning schema. As for tools, we have implemented all 37 tools using AgentLego \footnote{ \url{https://github.com/InternLM/agentlego}}, which is a platform supporting tool serving and remote accessing. When evaluating different LLMs, we replace different LLMs into the Lagent framework, and evaluate this system on the Opencompass \footnote{ \url{https://github.com/open-compass/opencompass}} evaluation platform.

\subsection{Prompts Used in GTA-Workflow}
\label{appendix:prompt}
The ReAct-style prompt template using for Lagent system is shown in Figure~\ref{fig:prompt}.
The raw query and checkpoint construction prompt of GTA-Workflow is shown in Figure~\ref{fig:workflow_query_generation}. 
The prompts of task classification, refinement and augmentation are shown in Figure~\ref{fig:workflow_classification}, \ref{fig:workflow_refinement}-\ref{fig:workflow_refinement_2}, and \ref{fig:workflow_augmentation}-\ref{fig:workflow_augmentation_2}.
The task validation and checkpoint regeneration prompts are shown in Figure~\ref{fig:workflow_rewrite} and \ref{fig:workflow_checkpoint_regenerate}, respectively.
To ensure the quality and realism of GTA-Workflow tasks, all tasks undergo manual quality inspection according to the criteria shown in Figure~\ref{fig:workflow_human}.
The prompt of LLM judge is shown in Figure~\ref{fig:workflow_judge}.
The prompt used for three-level failure decomposition is shown in Figure~\ref{fig:three_tier_classification}.

\subsection{Failure type description for harness failure analysis}
The detailed description of four stages of failures in Table~\ref{tab:harness_error_distribution} is listed below:
\begin{itemize}
    \item Data extraction: retrieval, verification, citation, source gathering, field extraction.
    \item Reasoning: recommendation quality, justification, rationale, trade-off discussion.
    \item Content synthesis: summarization, analysis, computation, synthesis, narrative composition.
    \item Formatting: file format, filename, layout, embedding, export, packaging, delivery artifact compliance.
\end{itemize}

\subsection{Task Examples}
\begin{figure*}[htbp]
    \centering
    \includegraphics[width=0.97\textwidth]{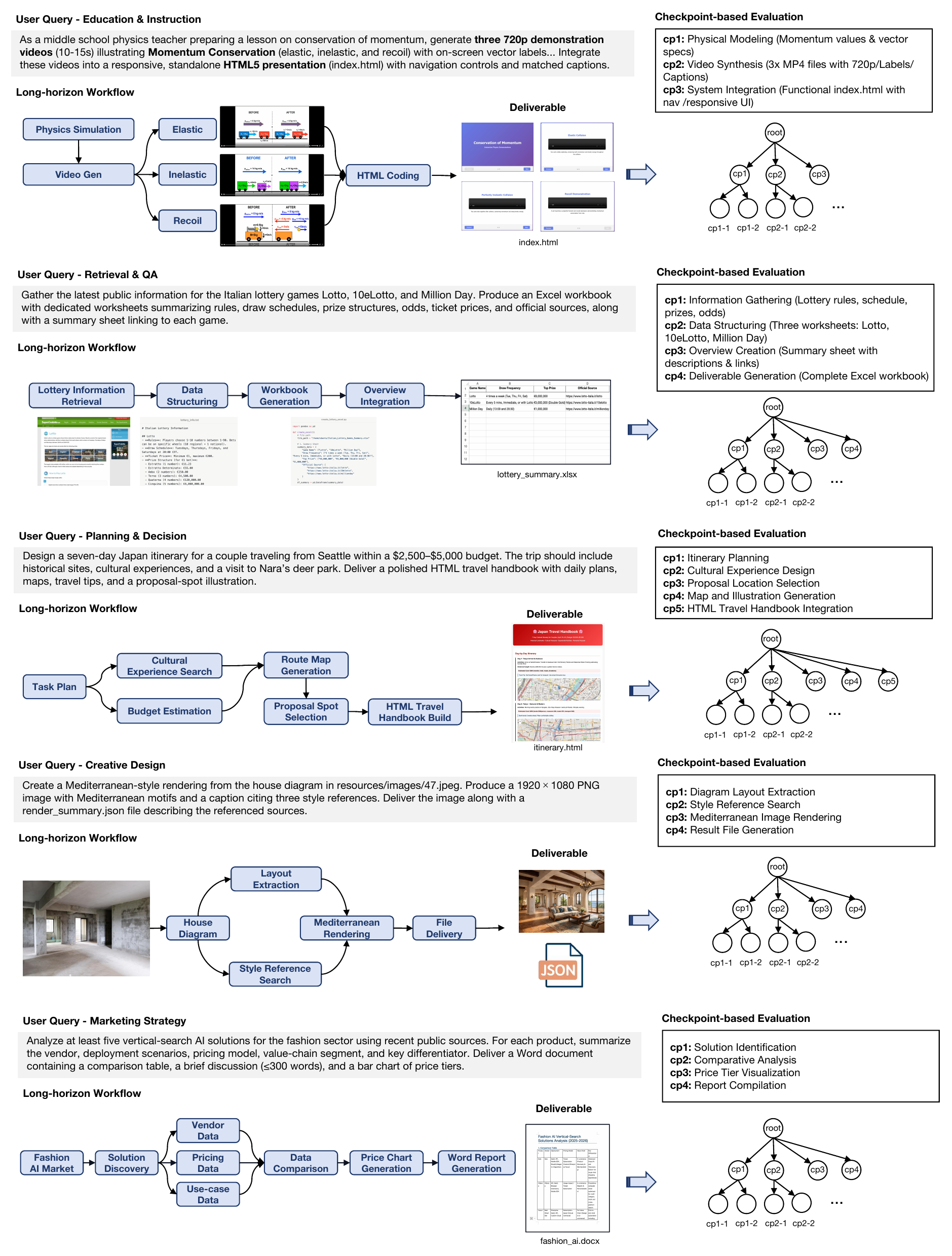}
    \caption{Task examples of GTA-Workflow from different categories. For presentation purpose, both the queries and checkpoints are condensed versions.}
    \label{fig:workflow_examples}
\end{figure*}

Task examples of GTA-Atomic are shown in Figure~\ref{fig:obj_query} to Figure~\ref{fig:img_query}. 
Task examples from 6 categories of GTA-Workflow are shown in Figure~\ref{fig:workflow_examples} and the lower part of Figure~\ref{fig:sample}.

\input{figures/tex_files/tool_def}
\input{figures/tex_files/tool_def_extension}
\input{figures/tex_files/prompt}

\input{figures/tex_files/workflow_query_generation}

\input{figures/tex_files/workflow_classification}

\input{figures/tex_files/workflow_refinement}

\input{figures/tex_files/workflow_augmentation}

\input{figures/tex_files/workflow_validation}

\input{figures/tex_files/workflow_human}

\input{figures/tex_files/workflow_judge}

\input{figures/tex_files/three_tier_classification}

\input{figures/tex_files/query_type}

\subsection{Details of GTA-Atomic Construction}
\label{sec:gta1}

\input{figures/tex_files/gta_atomic}

%% file: figures/tex_files/tool_def.tex
\begin{table*}[htbp]
\renewcommand{\arraystretch}{1.3} 
  \caption{Detailed definition of 14 tools across four categories in GTA-Atomic.}
  \label{tab:tool_desc}
  \centering
  \resizebox{0.95\textwidth}{!}{
  \begin{tabular}{l|p{5cm}|p{5cm}|p{5cm}}
    \toprule
\textbf{Name}  &\textbf{Description} & \textbf{Input} & \textbf{Output} \\
 \midrule
\multicolumn{4}{l}{{\cellcolor{gray!20}\textit{\textbf{- Perception}}}}\\
 OCR &Recognize the text from an image. &[image] An image containing text. &[text] The text on the image. \\
 RegionAttributeDesc. &Describe a certain attribute of a certain part in the input image.&[image] Any image. [text] Region location and the name of attribute to describe. &[text] The description of the region. \\
 DetectGivenObject&Detect certain object in the image. &[image] Any image. [text] Object name. &[image] An image with bounding box. [text] The location of bounding box and detecting scores. \\
 ImageDescription & Describe the input image.&[image] Any image. &[text] The description of the image. \\
\midrule
\multicolumn{4}{l}{{\cellcolor{gray!20}\textit{\textbf{- Operation}}}}\\
 DrawBox &Draw a box on a certain location of the image. &[image] Any image. [Text] Box location. &[image] An image with a box on the certain location. \\
 AddText &Add text on the image.&[image] Any image. [Text] Text, font size, and location. &[image] An image with text on the certain location. \\
 GoogleSearch &Search on Google.&[text] The content to search. &[text] Searching results. \\
\midrule
\multicolumn{4}{l}{{\cellcolor{gray!20}\textit{\textbf{- Logic}}}}\\
 Calculator &Calculate by Python interpreter.&[text] Math expressions including only numbers and operation symbols. &[text] Calculation result. \\
 Plot &Use code interpreter to draw math diagrams, statistics, etc.& [text] Python codes using Matplotlib to draw a diagram. & [image] The diagram. \\
 MathOCR &Recognize the math expressions from a image.&[image] An image containing math expression. & [text] Latex format of the math expression. \\
 CountGivenObject &Count the number of certain objects in the image.& [image] Any image. [text] The object name.& [text] The number of the object contained in the image.\\
 Solver &Use code interpreter to solve math expressions.&[text] Python codes using Sympy to solve math equations or expressions containing unknown variables. & [text] Solving results.\\
\midrule
\multicolumn{4}{l}{{\cellcolor{gray!20}\textit{\textbf{- Creativity}}}}\\
TextToImage &Generate an image from the input text.&[text] The description of an image. &[image] The image generated.  \\
 ImageStylization &Transfer the style of the image as that of a reference image.&[text] The description of the target image style. [image] An image to be transferred. &[image] The target image in the style of the text description. \\
\bottomrule
  \end{tabular}
  }%
\end{table*}

%% file: figures/tex_files/tool_def_extension.tex
\begin{table*}[htbp]
\renewcommand{\arraystretch}{1.3}
\caption{Extended tools in GTA-Workflow.}
\label{tab:ex_tool_desc}
\centering
\resizebox{0.95\textwidth}{!}{
\begin{tabular}{l|p{5cm}|p{5cm}|p{5cm}}
\toprule
\textbf{Name}  &\textbf{Description} & \textbf{Input} & \textbf{Output} \\
\midrule

\multicolumn{4}{l}{{\cellcolor{gray!20}\textit{\textbf{- Perception}}}}\\

ImageQualityTest &Evaluate image quality using Laplacian variance as a sharpness proxy. &
[image] The input image. &
[text] Image quality result including Laplacian variance and blur indicator. \\

SoundCharacteristicExtraction &Extract sound characteristics (MFCC, spectral centroid, spectral rolloff, zero crossing rate) from an audio file. &
[text] Path to the input audio file. &
[text] Extracted sound characteristics in JSON format. \\

VideoDescription &Generate a textual description of a video by captioning key frames. &
[text] Path to the input video file. &
[text] Generated video description. \\

VideoObjectDetection &Detect objects in video frames using a YOLO model. &
[text] Path to the input video file. &
[text] Object detection results for processed frames. \\

VideoOCR &Extract text appearing in video frames using OCR. &
[text] Path to the input video file. &
[text] Extracted text with bounding boxes. \\

SpeechToText &Convert speech in an audio file into text. &
[text] Path to the input audio file. &
[text] Transcribed text from the audio. \\

\midrule
\multicolumn{4}{l}{{\cellcolor{gray!20}\textit{\textbf{- Operation}}}}\\

CsvFileGenerator &Generate a CSV file from structured tabular data. &
[text] Filename. [text] JSON string representing a 2D array of data rows. &
[text] Absolute path of the generated CSV file. \\

DocxFileGenerator &Generate a DOCX file with the provided text content and optional images. &
[text] Filename. [text] Content with image placeholders. [text] Image paths. &
[text] Absolute path of the generated DOCX file. \\

PdfFileGenerator &Generate a PDF file with the provided text content and optional images. &
[text] Filename. [text] Content with image placeholders. [text] Image paths. &
[text] Absolute path of the generated PDF file. \\

PptxFileGenerator &Generate a PowerPoint presentation from structured slide content. &
[text] Filename. [text] JSON string describing slides. [text] Image paths. &
[text] Absolute path of the generated PPTX file. \\

XlsxFileGenerator &Generate an XLSX spreadsheet from structured tabular data. &
[text] Filename. [text] JSON string representing rows of data. &
[text] Absolute path of the generated XLSX file. \\

ReadCSV &Read a CSV file and return its content. &
[file] CSV file. &
[text] CSV content as a string. \\

ReadDOCX &Read a DOCX file and return its text content. &
[file] DOCX file. &
[text] Extracted text from the DOCX file. \\

ReadPDF &Read a PDF file and return its text content. &
[file] PDF file. &
[text] Extracted text from the PDF file. \\

ReadPPTX &Read a PPTX file and return its text content. &
[file] PPTX file. &
[text] Extracted text from the PPTX file. \\

ReadXLSX &Read an XLSX file and return its content as CSV-formatted text. &
[file] XLSX file. &
[text] Extracted XLSX content as CSV text. \\

AudioClipTool &Extract a clip from an audio file between specified start and end times. &
[text] Path to the input audio file. [int] Start time (optional). [int] End time (optional). &
[audio] Extracted audio clip. \\

NoiseReduction &Reduce noise in an audio file using spectral gating. &
[text] Path to the input audio file. &
[audio] Audio file with reduced noise. \\

PitchShifting &Shift the pitch of an audio file. &
[text] Path to the input audio file. &
[audio] Pitch-shifted audio file. \\

VideoClip &Extract a clip from a video between specified timestamps. &
[text] Path to the input video file. [int] Start time (optional). [int] End time (optional). &
[video] Extracted video clip. \\

AddTextVideoTool &Add text annotations or watermarks to a video. &
[text] Path to the input video file. &
[video] Video with text overlay. \\

HtmlFileGenerator &Generate an HTML file from provided markup or HTML code. &
[text] Output filename. [text] HTML markup content. &
[file] Generated HTML file path. \\

\midrule
\multicolumn{4}{l}{{\cellcolor{gray!20}\textit{\textbf{- Logic}}}}\\

Prover &Use the Z3 SMT solver to verify mathematical theorems. &
[text] SMT-LIB2 formatted logical formula string. &
[text] Proof result or counterexample. \\

\midrule
\multicolumn{4}{l}{{\cellcolor{gray!20}\textit{\textbf{- Creativity}}}}\\

TextToVideoTool &Generate a video from a text description using a diffusion model. &
[text] Video description. &
[video] Generated video. \\

\bottomrule
\end{tabular}
}
\end{table*}

%% file: figures/tex_files/prompt.tex
\begin{figure*}[htbp]
\centering
    \begin{tcolorbox}[colframe=black, colback=white]

\begin{verbatim}

CALL_PROTOCOL_EN = """You are a assistant who can utilize external tools. 
{tool_description}
To use a tool, please use the following format:
```
{thought}Think what you need to solve, do you need to use tools?
{action}the tool name, should be one of [{action_names}]
{action_input}the input to the action
```
The response after utilizing tools should using the following format:
```
{response}the results after call the tool.
```
If you already know the answer, or you do not need to use tools, please using the 
following format to reply:
```
{thought}the thought process to get the final answer
{finish}final answer
```
Begin!"""
\end{verbatim}
\end{tcolorbox}
\caption{The ReAct-style prompt template for the agent system.}
\label{fig:prompt}
\end{figure*}

%% file: figures/tex_files/workflow_query_generation.tex
\begin{figure*}[htbp]
\centering
    \begin{tcolorbox}[colframe=black, colback=white]

Please refer to the following structured example and generate data with the same structure for the user query: 

\texttt{\{EXAMPLE\_JSON\}}

All available tools are listed below: 

\texttt{\{TOOLS\_PROMPT\}}

\begin{itemize}
\item[1.] Based on the content of the query, generate result-oriented checkpoints organized in a tree-structured logical hierarchy with nested subtasks. The generated data is intended to evaluate the degree to which a large language model completes the task.

\item[2.] Follow the structure in the example strictly. The \texttt{tools} field must contain the tools required for solving the task, and tools can only be selected from the provided list. Each tool must include the fields \texttt{name}, \texttt{description}, \texttt{inputs}, and \texttt{outputs}. The \texttt{files} field refers to the files required as input to solve the query. The \texttt{dialogs} field represents the user query, and the \texttt{content} field must copy the original question exactly without any modification. The \texttt{sub\_tasks} field defines the checkpoint design for subtasks and must include \texttt{id}, \texttt{requirements} (the requirement of the subtask), \texttt{weight} (importance of the subtask, chosen from \{1, 2, 3\}), and \texttt{sub\_tasks} (nested subtasks if a hierarchical structure exists).

\item[3.] For simple problems, include 2--4 checkpoints. For medium-difficulty problems, include 5--7 checkpoints, typically with two levels of nesting. For difficult problems, include 8--10 checkpoints.

\item[4.] The output must be strictly in JSON format. Do not include any code block markers, comments, or extra text. All nested JSON strings must be processed using \texttt{json.dumps()}. All keys and values must use double quotes.
\end{itemize}

\end{tcolorbox}
\caption{The prompt of raw query and checkpoint generation in GTA-Workflow.}
\label{fig:workflow_query_generation}
\end{figure*}

%% file: figures/tex_files/workflow_classification.tex
\begin{figure*}[htbp]
\centering
    \begin{tcolorbox}[colframe=black, colback=white]

You are a data quality analyst for an AI agent benchmarking dataset. Your task is to classify each problem into one of four categories based on specific criteria.

\textbf{Classification Categories:}

\begin{itemize}
\item[1.] DELETE - Problems that should be removed entirely
\item[2.] REFINE - Problems that need requirement specification and output format clarification  
\item[3.] AUGMENT - Problems that need complexity expansion and tool augmentation
\item[4.] PASS - Problems that meet all quality standards and require no modification
\end{itemize}

\textbf{Classification Criteria:}

\textbf{Category 1: DELETE}
\begin{itemize}
\item[-] Problems requiring deep visual/video understanding beyond basic description (e.g., detailed scene analysis, complex object relationships)
\item[-] Problems that are essentially pure VLM (Vision-Language Model) evaluation tasks
\item[-] Problems where VideoDescription or ImageDescription would be insufficient for the required depth of analysis
\item[-] Problems that cannot be completed with the specified tools AND have no viable adaptation path using available tools while maintaining the core topic
\end{itemize}

\textbf{Category 2: REFINE}
\begin{itemize}

\item[-] Problems with vague, overly simple, or incomplete descriptions
\item[-] Problems lacking clear output format specification
\item[-] Problems with reasonable tool usage but unclear deliverables
\item[-] Problems that need clearer requirements and structured outputs
\end{itemize}

\textbf{Category 3: AUGMENT}
\begin{itemize}
\item[-] Problems involving webpage/app development (but can be redirected to requirements/documentation generation requiring diverse tools)
\item[-] Problems requiring no or minimal tool usage (e.g., general advice questions)
\item[-] Problems with only 1-2 tools that need expanded toolchains
\item[-] Problems that can be enhanced with multi-step workflows and diverse tools
\item[-] Problems that are too easy, containing less than 5 leaf node checkpoints
\item[-] Problems that cannot be completed with the specified tools BUT can be adapted to use available alternative tools OR can be modified while preserving the core topic to make it solvable
\end{itemize}

\textbf{Category 4: PASS}
\begin{itemize}
\item[-] Problems with clear, well-structured requirements
\item[-] Problems that specify concrete output formats (DOCX/PPTX/XLSX/PDF etc.)
\item[-] Problems with appropriate tool diversity (3-5 tools from multiple categories)
\item[-] Problems that represent realistic productivity scenarios
\item[-] Problems with proper complexity level (medium/hard difficulty)
\end{itemize}

\textbf{Tool Information:}
\texttt{\{tool\_categories\}}

\textbf{Task to Classify:}
\texttt{\{task\_content\}}

\textbf{Output Format:}
\begin{verbatim}
{
  "classification": "DELETE/REFINE/AUGMENT/PASS",
  "confidence": "high/medium/low",
  "reasoning": "Detailed explanation based on the criteria above",
  "specific_issues": ["List of specific problems identified (empty if PASS)"],
  "strengths": ["List of quality aspects that meet standards (for PASS category)"]
}

\end{verbatim}
\end{tcolorbox}
\caption{Query classification prompt in GTA-Workflow task construction.}
\label{fig:workflow_classification}
\end{figure*}

%% file: figures/tex_files/workflow_refinement.tex
\begin{figure*}[htbp]
\centering
    \begin{tcolorbox}[colframe=black, colback=white]

You are a requirements specification expert. Your task is to refine and improve the given problem by adding clarity, structure, and concrete deliverables. Besides, refine the checkpoint tree according to the improved problem.

\textbf{Refinement Goals:}
\begin{enumerate}
\item Add clear, specific requirements that eliminate ambiguity
\item Specify exact output format that could be generated by available content generation tools
\item Maintain the core intent while enhancing professionalism
\item Ensure the problem represents a realistic productivity scenarios
\end{enumerate}

\textbf{Refinement Guidelines:}
\begin{itemize}
\item Convert vague requests into specific, actionable tasks
\item Specify the exact output format from available generation tools
\item Include clear structure expectations (sections, visual elements, etc.)
\item Add measurable success criteria where possible
\item Ensure multi-step workflow is evident
\item Maintain natural, professional language
\end{itemize}

Avoid vague requirements descriptions, such as unclear timelines or ambiguous sources for information retrieval.  
Avoid including excessive descriptions of solution steps in the question (e.g. first, then, ...), and refrain from providing too many bullet points.  

You can only choose 1--2 tools from ``Input Parsing and Extraction'' category, 1--2 tools from ``Content Generation and Media Production'' category, and no limitation for the other two categories.  
Do not include new files as input resources that does not exist in the original task.

In addition to modifying the problem, the corresponding checkpoint must also be modified:

\textbf{Structured Checkpoint-based Evaluation}

\begin{itemize}
\item Each task must be decomposable into a hierarchical checkpoint tree structure
\item Checkpoints should be \textbf{result-oriented} with clearly defined success criteria
\item \textbf{Granularity requirements}:
    \begin{itemize}
    \item Leaf nodes must represent single, atomic actions that can be completed in one step
    \item Each leaf checkpoint must be objectively assessable by LLM judges
    \item Avoid vague or subjective evaluation criteria
    \end{itemize}

\item \textbf{Weighting system}:
    \begin{itemize}
    \item Assign appropriate weights (1--3) to each checkpoint based on importance
    \item Final score calculated through weighted aggregation from leaf nodes upward
    \item Parent nodes compute weighted averages of child node scores
    \end{itemize}

\item \textbf{Evaluation clarity}:
    \begin{itemize}
    \item Each checkpoint must have unambiguous completion criteria
    \item Focus on measurable outputs rather than process descriptions
    \item Ensure reliable and consistent scoring across different evaluators
    \end{itemize}
\end{itemize}

\textbf{Checkpoint Design Requirements:}

\begin{itemize}
\item 6--15 leaf node checkpoints per task depending on complexity 
      (6--8 for easy tasks, 9--12 for medium tasks, 13--15 for hard tasks)
\item Clear mapping between checkpoints and tool usage
\item Weighted scoring reflecting task importance
\item Leaf nodes must have objectively assessable outputs
\end{itemize}

\textbf{Tool Information:}

\texttt{\{tool\_categories\}}

\textbf{Original Task:}

\texttt{\{original\_task\}}

\end{tcolorbox}
\caption{Refinement prompt in GTA-Workflow task construction (1/2).}
\label{fig:workflow_refinement}
\end{figure*}

\begin{figure*}[htbp]
\centering
    \begin{tcolorbox}[colframe=black, colback=white]

\textbf{Specific Diagnosis of the Task by LLM Judge:}

In addition to the above guidelines, you may also refer to the LLM's diagnostic content for the original task when modifying it.

\texttt{\{diagnosis\}}

Your output should strictly adhere to the JSON structure of the original task when outputting the modified JSON format. Do not output any code block markers, comments, or extra text. Process all nested JSON strings using \texttt{json.dumps()}, and enclose all keys and values in double quotes. Do not include new files as input resources that does not exist in the original task.

\textbf{Output Format:}

\begin{verbatim}
{
  "refined_task": "Enhanced task with clear requirements and specified output 
  format. The task json format should be consistent with the original task 
  json format. Do not include new files as input resources that does not exist in 
  the original task.",
  "specified_deliverable": "e.g., DOCX report, PPTX presentation, XLSX analysis, 
  PDF document, CSV dataset, annotated images, processed audio/video",
  "content_generation_tool": "Specific tool required for output 
  (e.g., DocxFileGenerator, PptxFileGenerator, etc.)",
  "key_improvements": [
    "List of specific enhancements made",
    "e.g., 'Added output format specification'", 
    "e.g., 'Clarified analysis requirements'",
    "e.g., 'Specified visual/audio deliverables if applicable'"
  ],
  "complexity_level": "medium/hard"
}
\end{verbatim}

\end{tcolorbox}
\caption{Refinement prompt in GTA-Workflow task construction (2/2).}
\label{fig:workflow_refinement_2}
\end{figure*}

%% file: figures/tex_files/workflow_augmentation.tex
\begin{figure*}[htbp]
\centering
    \begin{tcolorbox}[colframe=black, colback=white]

You are a task complexity enhancement expert. Your task is to transform simple problems into complex, multi-step workflows that require diverse tool usage.

\textbf{Augmentation Goals:}
\begin{itemize}
\item[1.] Expand problem complexity to require 3 to 5+ tools total (3-4 for medium tasks, 5+ for hard tasks)
\item[2.] Ensure tool diversity across multiple categories
\item[3.] Create logical workflow with clear data dependencies
\item[4.] Specify concrete deliverables using content generation tools
\end{itemize}

\textbf{Tool Information}

\texttt{\{tool\_categories\}}

\textbf{Augmentation Rules:}
\begin{itemize}

\item[-] Web/App Development Problems: Redirect to requirement analysis, documentation, or prototyping
\item[-] Minimal Tool Problems: Expand to include input parsing, data processing, and content generation
\item[-] Tool Diversity: Include tools from at least 2 different categories. The current tool usage statistics are: \texttt{\{current\_tool\_usage\_stat\}}. You'd better choose low-frequency tools for tool diversity.
\item[-] Workflow Logic: Follow \textit{Input → Process → Enhance → Generate} sequence
\item[-] Underutilized Tools: Prioritize tools like Plot, Calculator, VideoObjectDetection, SoundCharacteristicExtraction, etc.
\item[-] GoogleSearch Limitation: Use only when external research is essential
\end{itemize}

If you need to rewrite the problem (since there may be newly added tools), there are some problem scenarios (but not limited to these) for your reference:
The problems should focus on Real-world Productivity Scenarios
\begin{itemize}
   \item[-] Focus on Intelligent Office / Life Productivity Assistant use cases
   \item[-] Target performing complex document / file processing and generation tasks
   \item[-] Scenarios include: 
        \textit{Professional Domains:}
        \begin{itemize}
        \item[•] Business analysis \& strategic planning
        \item[•] Research synthesis \& reporting
        \item[•] Data analysis \& visualization
        \item[•] Document creation \& presentation
        \item[•] More ...
        \end{itemize}
        \textit{Personal \& Creative Domains:}
        \begin{itemize}
        \item[•] Travel planning \& itinerary design
        \item[•] Learning projects \& research compilation
        \item[•] Content creation (video editing, blogging, social media)
        \item[•] Event planning \& personal organization
        \item[•] More ...
        \end{itemize}
        \textit{Core Task Characteristics:}
        \begin{itemize}
        \item[•] Multi-step workflows with clear progression
        \item[•] Information gathering → processing → output generation
        \item[•] Practical outputs with real-world utility
        \item[•] Professional-grade deliverables
        \end{itemize}
\end{itemize}

Avoid vague requirements descriptions, such as unclear timelines or ambiguous sources for information retrieval.
Avoid including excessive descriptions of solution steps in the question (e.g. first, then, ...), and refrain from providing too many bullet points. 
You can only choose 1-2 tools from "Input Parsing and Extraction" category, 1-2 tools from "Content Generation and Media Production" category, and no limitation for the other two categories.
Do not include new files as input resources that does not exist in the original task.

\end{tcolorbox}
\caption{Augmentation prompt in GTA-Workflow task construction (1/2).}
\label{fig:workflow_augmentation}
\end{figure*}

\begin{figure*}[htbp]
\centering
    \begin{tcolorbox}[colframe=black, colback=white]

In addition to modifying the problem, the corresponding checkpoint must also be modified:

    \textbf{Structured Checkpoint-based Evaluation}
    \begin{itemize}
    \item[-] Each task must be decomposable into a hierarchical checkpoint tree structure
    \item[-] Checkpoints should be result-oriented with clearly defined success criteria
    \item[-] Granularity requirements:
    \begin{itemize}
     \item[•] Leaf nodes must represent single, atomic actions that can be completed in one step
     \item[•] Each leaf checkpoint must be objectively assessable by LLM judges
     \item[•] Avoid vague or subjective evaluation criteria
     \end{itemize}
    \item[-] Weighting system:
    \begin{itemize}
     \item[•] Assign appropriate weights (1-3) to each checkpoint based on importance
     \item[•] Final score calculated through weighted aggregation from leaf nodes upward
     \item[•] Parent nodes compute weighted averages of child node scores
     \end{itemize}
    \item[-] Evaluation clarity:
    \begin{itemize}
     \item[•] Each checkpoint must have unambiguous completion criteria
     \item[•] Focus on measurable outputs rather than process descriptions
     \item[•] Ensure reliable and consistent scoring across different evaluators
     \end{itemize}
     \end{itemize}
     
     \textbf{Checkpoint Design Requirements:}
     \begin{itemize}
    \item[-] 6-15 leaf node checkpoints per task depending on complexity (6-8 for easy tasks, 9-12 for medium tasks, 13-15 for hard tasks)
    \item[-] Clear mapping between checkpoints and tool usage
    \item[-] Weighted scoring reflecting task importance
    \item[-] Leaf nodes must have objectively assessable outputs
    \end{itemize}
   
\textbf{Original Problem:}
\texttt{\{original\_problem\}}

\textbf{Specific Diagnosis of the Task by LLM Judge}
In addition to the above guidelines, you may also refer to the LLM's diagnostic content when modifying the task.
\texttt{\{diagnosis\}}

Your output should strictly adhere to the JSON structure of the original task when outputting the modified JSON format. Do not output any code block markers, comments, or extra text. Process all nested JSON strings using `json.dumps()`, and enclose all keys and values in double quotes. Do not include new files as input resources that does not exist in the original task.

\textbf{Output Format:}
\begin{verbatim}
{
  "augmented_task": "Complex problem with multi-step workflow and clear 
  deliverables. The task json format should be consistent with the original task 
  json format. Do not include new files as input resources that does not exist in 
  the original task.",
  "recommended_tools": ["tool1", "tool2", "tool3", "tool4", "tool5"],
  "tool_categories_covered": ["input_parsing", "data_processing", 
  "information_enhancement", "content_generation"],
  "workflow_description": "Step-by-step explanation of tool sequence and data 
  flow",
  "specified_deliverable": "Concrete output format and content expectations",
  "complexity_level": "medium/hard"
}

\end{verbatim}
\end{tcolorbox}
\caption{Augmentation prompt in GTA-Workflow task construction (2/2).}
\label{fig:workflow_augmentation_2}
\end{figure*}

%% file: figures/tex_files/workflow_validation.tex
\begin{figure*}[htbp]
\centering
    \begin{tcolorbox}[colframe=black, colback=white]

You will rewrite a task description to remove explicit tool-call phrasing and convert enumerated lists into coherent narrative text while preserving requirements and deliverables.

\textbf{Rewrite Rules}

\begin{itemize}

\item Remove or rephrase any explicit mentions such as ``use X tool'', ``call Y'', or ``via Z tool''. Describe required outcomes neutrally (e.g., ``extract'', ``analyze'', ``generate'') without naming tools in the instruction text.

\item Convert numbered or bulleted requirements into clear, cohesive paragraphs while keeping all constraints, quantities, success criteria, and deliverables intact.

\item Preserve the task's meaning, scope, outputs, and formatting requirements.

\item Do not introduce or remove input files or change their names.

\item Do not add new tools or files.

\item Keep the values of the fields for \texttt{files} and any tool list unchanged.

\item Write in natural, flowing prose suitable for a task brief; avoid list formatting and enumerations such as (1), 1., bullets, or stepwise scaffolding such as ``first'', ``then'', or ``finally''. Express the same requirements as connected narrative.

\end{itemize}

\textbf{Input JSON}

\begin{verbatim}
{
    "field_name": "refined_task | augmented_task",
    "text": <ORIGINAL_TASK_TEXT>,
    "files": <FILES_ARRAY>,
    "content_generation_tool": <CONTENT_GENERATION_TOOL or null>,
    "recommended_tools": <RECOMMENDED_TOOLS or null>
}
\end{verbatim}

\textbf{Output Format}

Output strictly valid JSON with exactly the following keys depending on the value of \texttt{field\_name}:

\begin{itemize}

\item \begin{verbatim}
If field_name == "refined_task":
{"refined_task": "...", "files": [...], "recommended_tools": [...]}
\end{verbatim}

\item \begin{verbatim}
If field_name == "augmented_task":
{"augmented_task": "...", "files": [...], "recommended_tools": [...]}
\end{verbatim}

\end{itemize}

Return only JSON. No markdown, no commentary.

\end{tcolorbox}
\caption{Task rewriting prompt of GTA-Workflow.}
\label{fig:workflow_rewrite}
\end{figure*}

\begin{figure*}[htbp]
\centering
    \begin{tcolorbox}[colframe=black, colback=white]

\textbf{General Goal:} 

Your task is to generate the \textbf{final \texttt{JSON}-formatted task object} with a complete hierarchy of \textbf{result-oriented checkpoints}, following exactly the schema below.

You will be given:

\begin{itemize}
\item The rewritten user query (from Step 2)
\item The original \texttt{JSON} structure (fields must be preserved)
\end{itemize}

Your job:

\begin{enumerate}
\item Insert the rewritten query into the \texttt{JSON} under the correct field.
\item Generate a complete set of \textbf{result-oriented} checkpoints in the \texttt{'sub\_tasks'} field.
\item Output strictly valid \texttt{JSON} and nothing else.
\end{enumerate}

\end{tcolorbox}
\caption{Checkpoint regeneration prompt of GTA-Workflow.}
\label{fig:workflow_checkpoint_regenerate}
\end{figure*}

%% file: figures/tex_files/workflow_human.tex
\begin{figure*}[htbp]
\centering
    \begin{tcolorbox}[colframe=black, colback=white]

\textbf{1. Task-level Quality Control}

The inspection must be conducted in conjunction with the task attachments and input files.

\begin{itemize}

\item \textbf{Clarity and Unambiguity}

\begin{itemize}
\item \textbf{Standard}: The task objective must be clearly stated and free from ambiguity. An independent reader should be able to understand what needs to be accomplished without confusion regarding the expected final deliverable.

\item \textbf{Checkpoints}:
\begin{itemize}
\item Are there ambiguous references or unclear pronouns?
\item Is the objective sufficiently specific? For example, a vague request such as ``find a good restaurant'' should be refined into a concrete requirement such as ``find a Chinese restaurant near location X with a rating above 4.5 and an average cost around Y per person.''
\end{itemize}
\end{itemize}

\item \textbf{Realism and Practicality}

\begin{itemize}
\item \textbf{Standard}: The task must simulate realistic requests that a human might reasonably pose in real-world scenarios. Tasks should not be artificially constructed, meaningless, or overly theoretical.

\item \textbf{Checkpoints}:
\begin{itemize}
\item Would this task realistically appear in real-life or professional settings?
\item Is the task background and contextual information reasonable and coherent?
\end{itemize}
\end{itemize}

\item \textbf{Tool Solvability}

\begin{itemize}
\item \textbf{Standard}: The task must be solvable using the toolset provided in the benchmark platform. Tasks requiring capabilities outside the available tools are not allowed.

\item \textbf{Checkpoints}:
\begin{itemize}
\item Verify the predefined tool chain from start to finish.
\item Ensure that each step can be executed by the available tools and that the final solution can be obtained through these tools.
\end{itemize}
\end{itemize}

\end{itemize}

\textbf{2. Workflow-specific Quality Control}

\begin{itemize}

\item \textbf{Workflow Structure Validity}

\begin{itemize}
\item \textbf{Standard}: The task must contain a non-linear workflow structure requiring dynamic planning (e.g., branching or conditional decisions), or consist of multiple complex and interdependent subtasks.

\item \textbf{Checkpoints}:
\begin{itemize}
\item \textbf{Branching}: Does the task require decision-making based on intermediate results? 
For example, if the weather query indicates rain, the agent should plan indoor activities; if it is sunny, outdoor activities should be planned.

\item \textbf{Dependency}: Do later subtasks strictly depend on the outputs of preceding subtasks?
\end{itemize}

\end{itemize}

\item \textbf{Checkpoint Definition Quality}

\begin{itemize}
\item \textbf{Standard}: Each checkpoint must correspond to a concrete and verifiable intermediate state or deliverable.

\item \textbf{Checkpoints}:
\begin{itemize}
\item Is the checkpoint description clear and specific? 
For example, ``successfully retrieved and summarized the key ideas of three relevant papers'' is verifiable, whereas ``did some research'' is not.

\item Can an evaluator objectively determine whether the agent has reached this stage solely based on the checkpoint description?
\end{itemize}

\end{itemize}

\end{itemize}

\textbf{3. Answer and Tool-chain Quality Control}

\begin{itemize}

\item \textbf{Evaluation Feasibility}

\begin{itemize}
\item \textbf{Standard}: The final output and intermediate checkpoints must be reliably assessable using the predefined automatic or human evaluation metrics.

\item \textbf{Checkpoints}:
\begin{itemize}
\item Given the final output, can evaluators consistently determine whether the result is correct or satisfactory?
\end{itemize}

\end{itemize}

\end{itemize}

\end{tcolorbox}
\caption{Human quality control guidelines for GTA-Workflow tasks.}
\label{fig:workflow_human}
\end{figure*}

%% file: figures/tex_files/workflow_judge.tex
\begin{figure*}[htbp]
\centering
    \begin{tcolorbox}[colframe=black, colback=white]

You are a professional LLM Agent evaluator. All of the generated files and images from the LLM Agent are uploaded, videos are converted to several image frames, and audio files are checked for existence.

\textbf{Original task:}

\texttt{\{task\}}

\textbf{Final answer:}

\texttt{\{final\_answer\}}

\textbf{Checkpoint requirement:}

\texttt{\{req\}}

Please score this checkpoint from 0 to 10. If required files are missing, give a low score.

Only output valid JSON:

\begin{verbatim}
{"score": number, "analysis": "short explanation (<80 words)"}
\end{verbatim}

\end{tcolorbox}
\caption{The prompt for LLM judge in GTA-Workflow.}
\label{fig:workflow_judge}
\end{figure*}

%% file: figures/tex_files/three_tier_classification.tex
\begin{figure*}[htbp]
\centering
    \begin{tcolorbox}[colframe=black, colback=white]

You are classifying one checkpoint by failure depth tier.

\textbf{Allowed tiers and labels:}
\begin{itemize}
    \item \textbf{A: Leaf-level failure} (local sub-goal not completed; e.g., incorrect extraction, missing section, or failed substep)
    \item \textbf{B: Mid-level composition failure} (sub-goals are mostly completed, but integration or coordination across components fails; e.g., inconsistent structure or incoherent aggregation)
    \item \textbf{C: Root-level deliverable failure} (final output fails in packaging, formatting, or submission compliance; e.g., incorrect file format, missing export, or invalid structure)
\end{itemize}

\textbf{Guidelines:}
\begin{itemize}
    \item Use the requirement text as the primary signal.
    \item Use structural context (leaf status, depth, parent) as auxiliary information.
    \item If the requirement mainly concerns final handoff, file export, formatting, filename, or submission compliance, prefer \textbf{C}.
    \item If the requirement mainly concerns combining multiple completed parts into a coherent module, page, report, or workflow, prefer \textbf{B}.
    \item If the requirement is atomic or local (e.g., a single extraction, check, section, or substep), prefer \textbf{A}.
    \item A non-leaf node can still be classified as \textbf{A} if its requirement is semantically local.
\end{itemize}

\textbf{Output format:} Return a JSON object with the following fields:
\begin{itemize}
    \item \texttt{tier}: one of \{A, B, C\}
    \item \texttt{tier\_label}: exact string matching the selected tier
    \item \texttt{confidence}: a float in $[0,1]$
    \item \texttt{reason}: a concise explanation (no more than 60 words)
\end{itemize}

\end{tcolorbox}
\caption{The classification prompt used for three-level failure decomposition.}
\label{fig:three_tier_classification}
\end{figure*}

%% file: figures/tex_files/query_type.tex
\begin{figure*}[htbp]
    \centering
    \begin{tcolorbox}[colframe=black, colback=white]

    \textbf{Query Type:} Objective
    
    \textbf{Query:}
    I need to prepare twelve servings of this dish. How many boxes of eggs will I need in total? 
    
    \textbf{Involved Tools:} ImageDescription, CountGivenObject, OCR 
    
    \begin{minipage}[c]{0.3\textwidth}
    \vspace{5pt}
    \raggedright
    \textbf{Files:}\\
    \vspace{2pt}
    \centering
    \includegraphics[bb=0 0 1280 1280,width=0.99\textwidth]{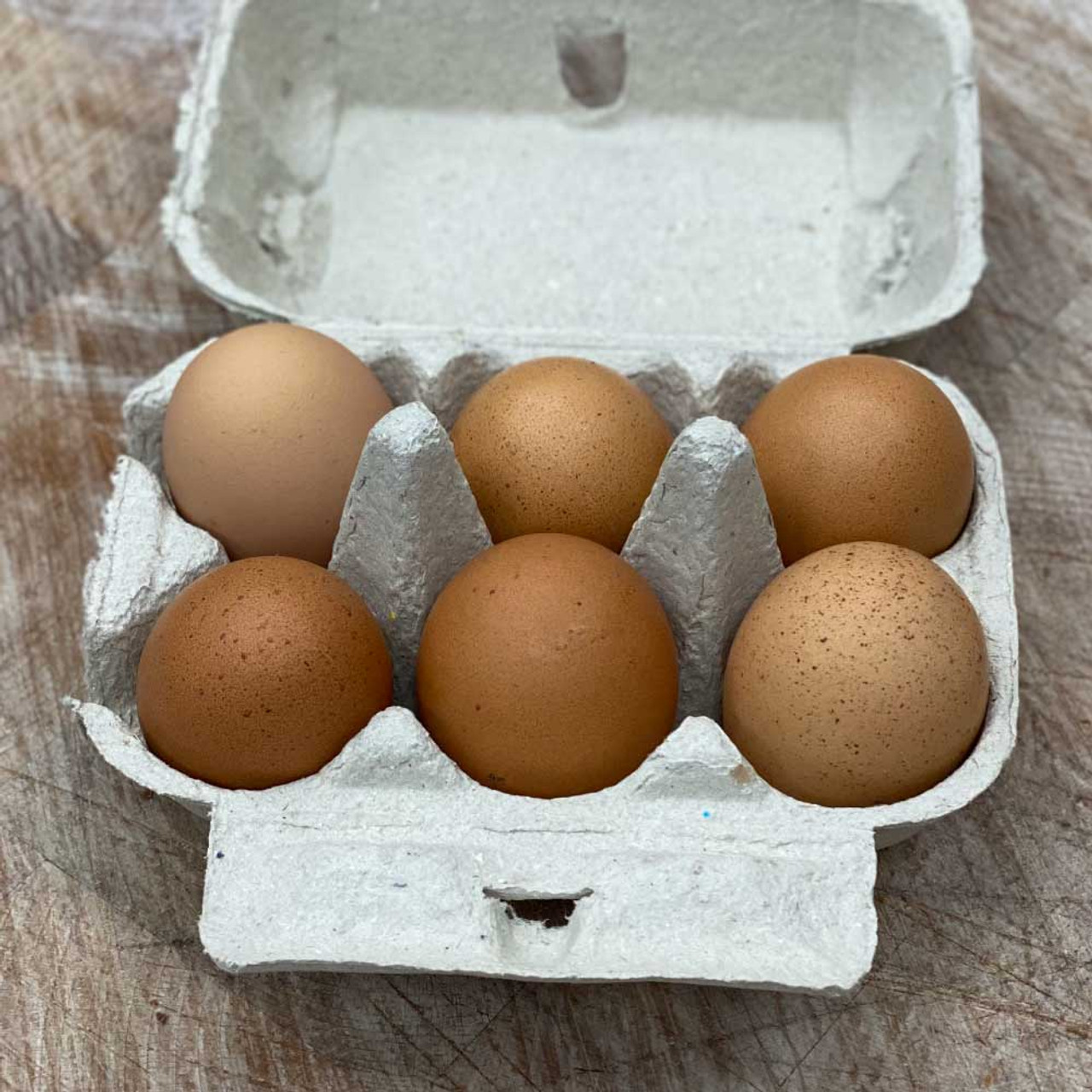}
    \end{minipage}
    \begin{minipage}[c]{0.69\textwidth}
    \vspace{16pt}
    \centering
    \includegraphics[width=0.99\textwidth]{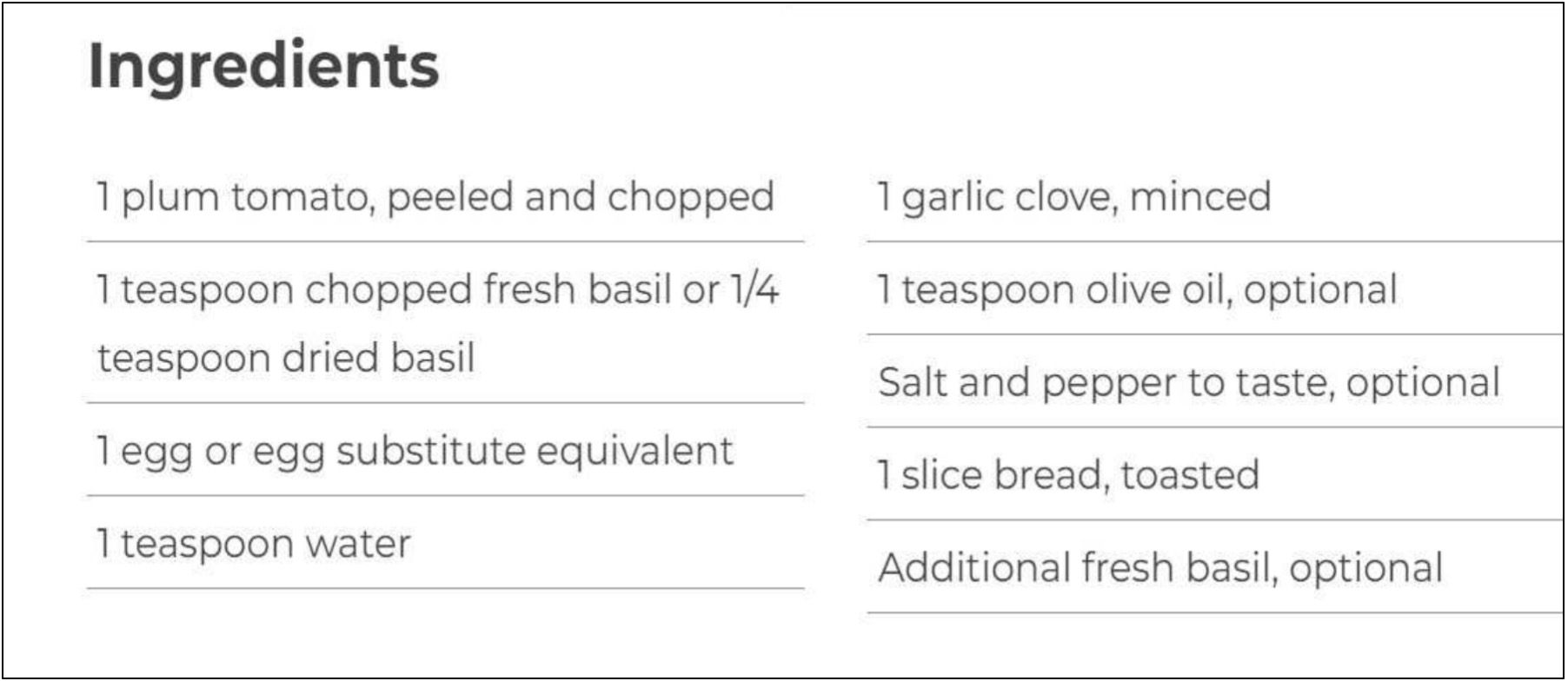}
    \end{minipage}
    
    \vspace{5pt}
    \textbf{Steps:} 
    \begin{itemize}
        \item[1.] Count the number of eggs in the photo.
        \item[2.] Identify the eggs needed for one serving of a dish on the recipe.
        \item[3.] Calculate how many eggs are needed for 12 dishes.
        \item[4.] Calculate how many boxes of eggs are needed.
    \end{itemize} 
    \textbf{Answer:} 2 

    \end{tcolorbox}
    \caption{An example of objective query in GTA-Atomic. The final answer is a uniquely determined number or phrase.}
    \label{fig:obj_query}
\end{figure*}

\begin{figure*}[htbp]
    \centering
    \begin{tcolorbox}[colframe=black, colback=white]

    \textbf{Query Type:} Subjective
    
    \textbf{Query:}
    According to the sign, what should I avoid to do now? Why?
    
    \textbf{Involved Tools:} ImageDescription, OCR 
    
    \begin{minipage}[c]{0.3\textwidth}
    \raggedright
    \textbf{Files:}\\
    \vspace{2pt}
    \centering
    \includegraphics[width=0.99\textwidth]{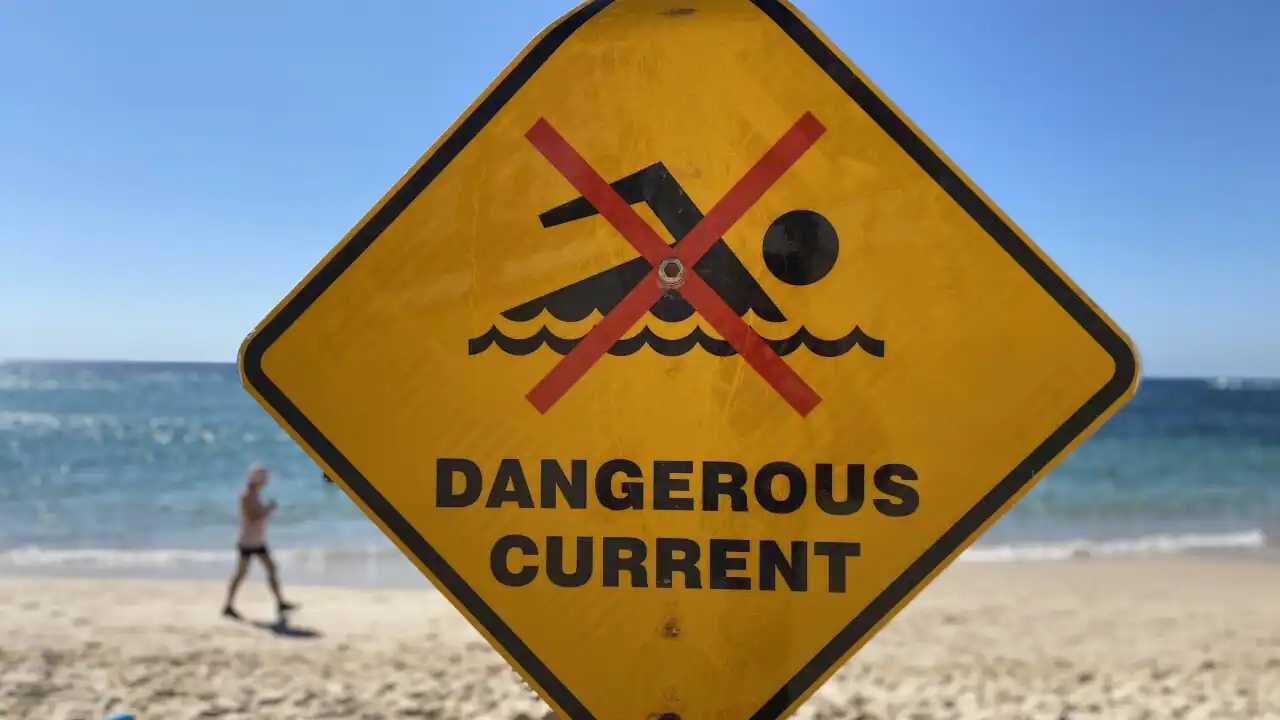}
    \end{minipage}
    \hfill
    \begin{minipage}[c]{0.66\textwidth}
    \vspace{6pt}
    \textbf{Steps:} 
    \begin{itemize}
        \item[1.] Recognize the image background and the icon on the sign.
        \item[2.] Recognize the text in the picture.
    \end{itemize} 
    \textbf{Answer:} You should avoid swimming due to the dangerous current.
    \end{minipage}
    
    \end{tcolorbox}
    \caption{An example of subjective query in GTA-Atomic. The final answer is usually some descriptive text. It is not unique, but the general idea is the same.}
    \label{fig:sbj_query}
\end{figure*}

\begin{figure*}[htbp]
    \centering
    \begin{tcolorbox}[colframe=black, colback=white]

    \textbf{Query Type:} Image Generation
    
    \textbf{Query:}
    I want to go to the highest-rated restaurant. Please circle it in the map.
    
    \textbf{Involved Tools:} OCR, DrawBox
    
    \begin{minipage}[c]{0.49\textwidth}
    \vspace{5pt}
    \raggedright
    \textbf{Files:}\\
    \vspace{2pt}
    \centering
    \includegraphics[width=0.99\textwidth]{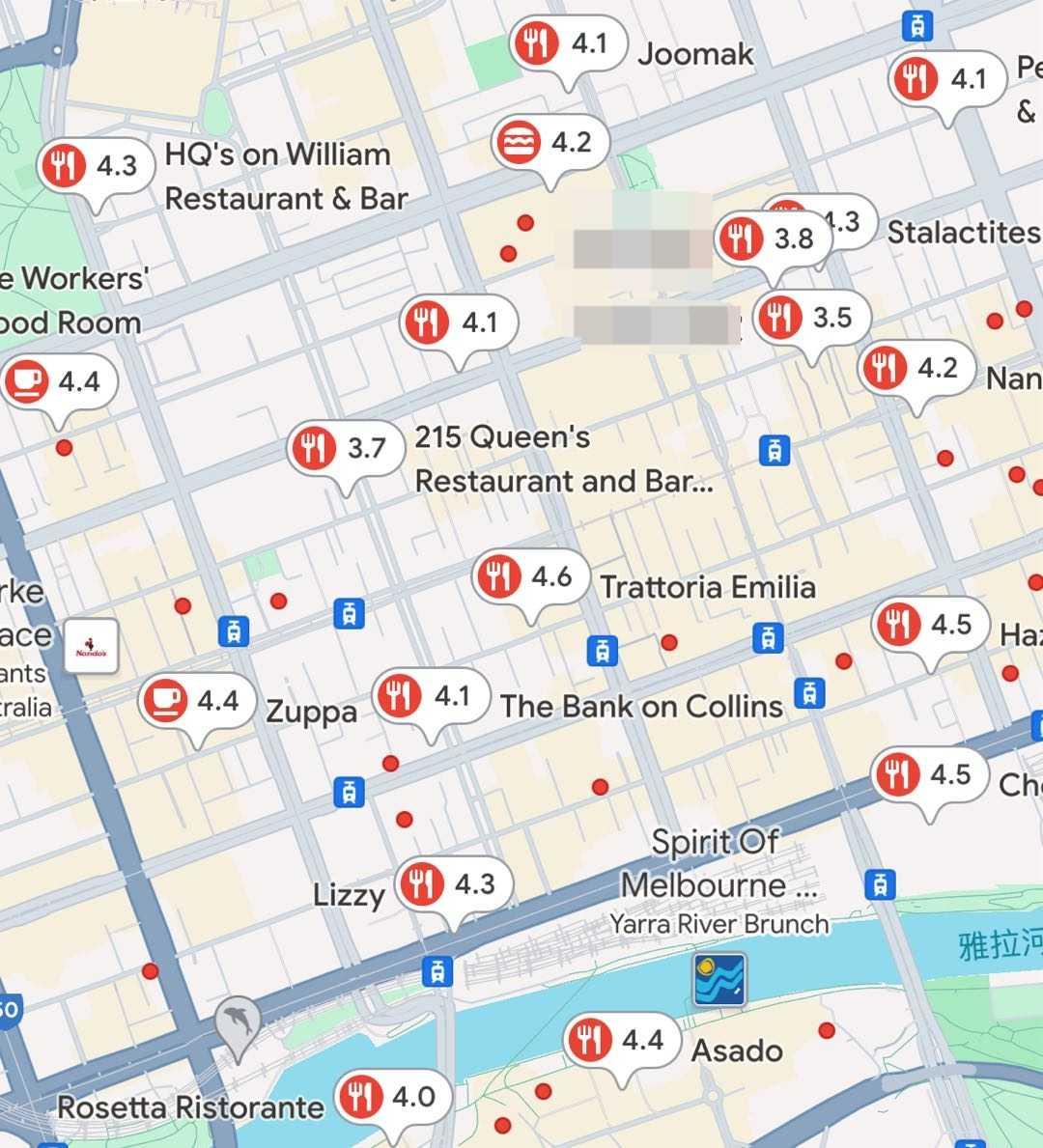}
    \end{minipage}
    \hfill
    \begin{minipage}[c]{0.49\textwidth}
    \vspace{5pt}
    \raggedright
    \textbf{Generated Image:} \\
    \vspace{2pt}
    \includegraphics[width=0.99\textwidth]{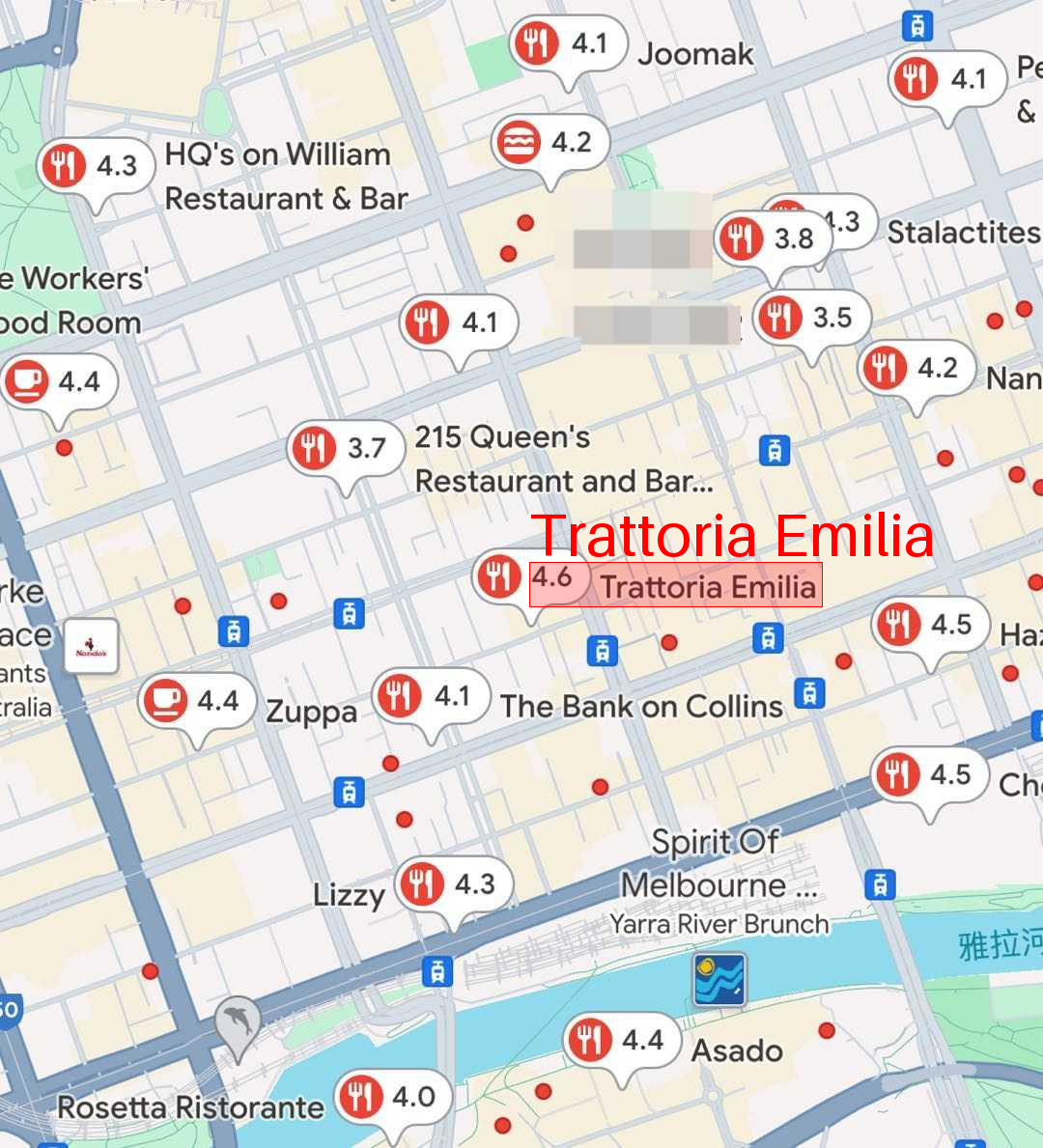}
    \end{minipage}

    \vspace{5pt}
    \textbf{Steps:} 
    \begin{itemize}
        \item[1.] Identify the ratings of each restaurant in the map using OCR tool.
        \item[2.] Identify the restaurant with the highest rating and its coordinate from the OCR result.
        \item[3.] Circle the restaurant in the graph using DrawBox tool.
    \end{itemize} 
    \end{tcolorbox}
    \caption{An example of image generation query in GTA-Atomic. The final answer is none since we do not evaluate the generated image directly.}
    \label{fig:img_query}
\end{figure*}

%% file: figures/tex_files/gta_atomic.tex
GTA-Atomic evaluates short-horizon, closed-ended tool-use tasks in realistic settings. 
Given a tool set $\mathcal{T}_c$, each sample is defined as $(\mathcal{F}, \mathcal{Q}, \mathcal{T}, \mathcal{C}, \mathcal{A})$, where $\mathcal{F}$ denotes input files (typically images), $\mathcal{Q}$ is a real-world query, $\mathcal{T} \subseteq \mathcal{T}_c$ is the set of involved tools, $\mathcal{C}$ is a multi-step reference tool chain, and $\mathcal{A}$ is the final answer. 
The tool chain $\mathcal{C}=\{(t_i, a_i, r_i)\}_{i=1}^m$ records step-wise tool invocations, including tool selection, input arguments, and outputs. 
Importantly, the query does not explicitly specify the required tools or steps, requiring models to perform reasoning and planning for tool use.

The tool set consists of 14 executable tools spanning four categories: perception, operation, logic, and creativity. 
Queries are categorized into objective, subjective, and image generation types. 
Objective queries have unique answers, while subjective queries allow multiple valid responses with consistent semantics. 
For image generation queries, evaluation focuses on tool invocation correctness, rather than the generated content.

To construct the dataset, we adopt an exemplar-based expansion pipeline, as shown in the upper part of Figure~\ref{fig:construct}. 
We first design a set of seed queries covering diverse real-world scenarios and tool combinations. 
Annotators then expand these exemplars by generating new queries with similar tool requirements but varied contexts, ensuring both diversity and controllability. 
All queries are required to be realistic, solvable using the provided tools, and free of explicit references to specific tools, so that tool selection must be inferred.

For each query, annotators manually construct the corresponding tool chain and final answer following a ReAct-style interaction format. 
They execute the tools step by step, record intermediate results, and ensure that each step is executable and logically consistent. 
Samples with incorrect tool behavior or ambiguous answers are discarded to ensure quality.

Overall, GTA-Atomic provides a high-fidelity benchmark for evaluating fine-grained tool-use precision, reasoning, and multi-step coordination, forming the foundation of the GTA-2 hierarchical framework.